\documentclass{article}
\usepackage[T1]{fontenc}
\usepackage[utf8]{inputenc}
\usepackage{amsmath,amssymb,amsfonts,mathrsfs,bm}
\usepackage{mathtools}
\usepackage{amsthm}
\usepackage{scalerel}
\usepackage{nicefrac}
\usepackage{microtype} 
\usepackage[shortlabels]{enumitem}
\usepackage{graphicx}
\usepackage{epstopdf}
\DeclareGraphicsExtensions{.eps,.png,.jpg,.pdf}

\usepackage{url}
\usepackage{colortbl}
\usepackage{booktabs}
\usepackage{multirow}
\usepackage[table,dvipsnames]{xcolor}
\usepackage[normalem]{ulem}
\usepackage{xparse}
\usepackage{calc}
\usepackage{etoolbox}

\makeatletter
\@ifpackageloaded{natbib}{
	\relax
}{
	\usepackage{cite}
}
\makeatother

\usepackage[small]{caption}

\usepackage{array}
\newcolumntype{L}[1]{>{\raggedright\let\newline\\\arraybackslash\hspace{0pt}}m{#1}}
\newcolumntype{C}[1]{>{\centering\let\newline\\\arraybackslash\hspace{0pt}}m{#1}}
\newcolumntype{R}[1]{>{\raggedleft\let\newline\\\arraybackslash\hspace{0pt}}m{#1}}

\makeatletter
\let\MYcaption\@makecaption
\makeatother
\usepackage[font=footnotesize]{subcaption}
\makeatletter
\let\@makecaption\MYcaption
\makeatother

\usepackage{glossaries}
\makeatletter
\sfcode`\.1006

\let\oldgls\gls
\let\oldglspl\glspl

\newcommand\fussy@ifnextchar[3]{%
	\let\reserved@d=#1%
	\def\reserved@a{#2}%
	\def\reserved@b{#3}%
	\futurelet\@let@token\fussy@ifnch}
\def\fussy@ifnch{%
	\ifx\@let@token\reserved@d
		\let\reserved@c\reserved@a
	\else
		\let\reserved@c\reserved@b
	\fi
	\reserved@c}

\renewcommand{\gls}[1]{%
\oldgls{#1}\fussy@ifnextchar.{\@checkperiod}{\@}}
\renewcommand{\glspl}[1]{%
\oldglspl{#1}\fussy@ifnextchar.{\@checkperiod}{\@}}

\newcommand{\@checkperiod}[1]{%
	\ifnum\sfcode`\.=\spacefactor\else#1\fi
}

\robustify{\gls}
\robustify{\glspl}
\makeatother

\newacronym{wrt}{w.r.t.}{with respect to}
\newacronym{RHS}{R.H.S.}{right-hand side}
\newacronym{LHS}{L.H.S.}{left-hand side}
\newacronym{iid}{i.i.d.}{independent and identically distributed}

\usepackage{float}
\usepackage[hidelinks]{hyperref}

\usepackage[capitalize]{cleveref}
\crefname{equation}{}{}
\Crefname{equation}{}{}
\crefname{claim}{claim}{claims}
\crefname{step}{step}{steps}
\crefname{line}{line}{lines}
\crefname{condition}{condition}{conditions}
\crefname{dmath}{}{}
\crefname{dseries}{}{}
\crefname{dgroup}{}{}

\crefname{Problem}{Problem}{Problems}
\crefformat{Problem}{Problem~(#2#1#3)}
\crefrangeformat{Problem}{Problems~(#3#1#4) to~(#5#2#6)}

\crefname{Theorem}{Theorem}{Theorems}
\crefname{Corollary}{Corollary}{Corollaries}
\crefname{Proposition}{Proposition}{Propositions}
\crefname{Lemma}{Lemma}{Lemmas}
\crefname{Definition}{Definition}{Definitions}
\crefname{Example}{Example}{Examples}
\crefname{Assumption}{Assumption}{Assumptions}
\crefname{Remark}{Remark}{Remarks}
\crefname{Rem}{Remark}{Remarks}
\crefname{remarks}{Remarks}{Remarks}
\crefname{Appendix}{Appendix}{Appendices}
\crefname{Supplement}{Supplement}{Supplements}
\crefname{Exercise}{Exercise}{Exercises}
\crefname{Theorem_A}{Theorem}{Theorems}
\crefname{Corollary_A}{Corollary}{Corollaries}
\crefname{Proposition_A}{Proposition}{Propositions}
\crefname{Lemma_A}{Lemma}{Lemmas}
\crefname{Definition_A}{Definition}{Definitions}

\usepackage{crossreftools}

\usepackage{algorithm}
\usepackage{algpseudocode}

\ifx\loadbreqn\undefined
	\relax
\else
	\usepackage{breqn}
\fi

\makeatletter
\def\cleartheorem#1{%
    \expandafter\let\csname#1\endcsname\relax
    \expandafter\let\csname c@#1\endcsname\relax
}
\def\clearthms#1{ \@for\tname:=#1\do{\cleartheorem\tname} }
\makeatother

\ifx\renewtheorem\undefined
	\ifx\useTheoremCounter\undefined
		\newtheorem{Theorem}{Theorem}
		\newtheorem{Corollary}{Corollary}
		\newtheorem{Proposition}{Proposition}
		
	\else

	\fi

\fi

\theoremstyle{remark}

\theoremstyle{plain}

\newcommand{\qednew}{\nobreak \ifvmode \relax \else
		\ifdim\lastskip<1.5em \hskip-\lastskip
			\hskip1.5em plus0em minus0.5em \fi \nobreak
		\vrule height0.75em width0.5em depth0.25em\fi}



\newcommand{\nn}{\nonumber\\ }

\NewDocumentCommand{\movedownsub}{e{^_}}{%
	\IfNoValueTF{#1}{%
		\IfNoValueF{#2}{^{}}
	}{%
		^{#1}
	}%
	\IfNoValueF{#2}{_{#2}}
}

\let\latexchi\chi
\RenewDocumentCommand{\chi}{}{\latexchi\movedownsub}

\newcommand{\Real}{\mathbb{R}}

\newcommand{\calE}{\mathcal{E}}
\newcommand{\calF}{\mathcal{F}}
\newcommand{\calG}{\mathcal{G}}

\newcommand{\calV}{\mathcal{V}}

\newcommand{\ba}{\mathbf{a}}
\newcommand{\bA}{\mathbf{A}}

\newcommand{\bv}{\mathbf{v}}
\newcommand{\bV}{\mathbf{V}}

\newcommand{\bW}{\mathbf{W}}
\newcommand{\bx}{\mathbf{x}}
\newcommand{\bX}{\mathbf{X}}

\DeclareSymbolFont{bsfletters}{OT1}{cmss}{bx}{n}
\DeclareSymbolFont{ssfletters}{OT1}{cmss}{m}{n}
\DeclareMathSymbol{\bsfGamma}{0}{bsfletters}{'000}
\DeclareMathSymbol{\ssfGamma}{0}{ssfletters}{'000}
\DeclareMathSymbol{\bsfDelta}{0}{bsfletters}{'001}
\DeclareMathSymbol{\ssfDelta}{0}{ssfletters}{'001}
\DeclareMathSymbol{\bsfTheta}{0}{bsfletters}{'002}
\DeclareMathSymbol{\ssfTheta}{0}{ssfletters}{'002}
\DeclareMathSymbol{\bsfLambda}{0}{bsfletters}{'003}
\DeclareMathSymbol{\ssfLambda}{0}{ssfletters}{'003}
\DeclareMathSymbol{\bsfXi}{0}{bsfletters}{'004}
\DeclareMathSymbol{\ssfXi}{0}{ssfletters}{'004}
\DeclareMathSymbol{\bsfPi}{0}{bsfletters}{'005}
\DeclareMathSymbol{\ssfPi}{0}{ssfletters}{'005}
\DeclareMathSymbol{\bsfSigma}{0}{bsfletters}{'006}
\DeclareMathSymbol{\ssfSigma}{0}{ssfletters}{'006}
\DeclareMathSymbol{\bsfUpsilon}{0}{bsfletters}{'007}
\DeclareMathSymbol{\ssfUpsilon}{0}{ssfletters}{'007}
\DeclareMathSymbol{\bsfPhi}{0}{bsfletters}{'010}
\DeclareMathSymbol{\ssfPhi}{0}{ssfletters}{'010}
\DeclareMathSymbol{\bsfPsi}{0}{bsfletters}{'011}
\DeclareMathSymbol{\ssfPsi}{0}{ssfletters}{'011}
\DeclareMathSymbol{\bsfOmega}{0}{bsfletters}{'012}
\DeclareMathSymbol{\ssfOmega}{0}{ssfletters}{'012}

\makeatletter
\newcommand*\rel@kern[1]{\kern#1\dimexpr\macc@kerna}
\newcommand*\widebar[1]{%
  \begingroup
  \def\mathaccent##1##2{%
    \rel@kern{0.8}%
    \overline{\rel@kern{-0.8}\macc@nucleus\rel@kern{0.2}}%
    \rel@kern{-0.2}%
  }%
  \macc@depth\@ne
  \let\math@bgroup\@empty \let\math@egroup\macc@set@skewchar
  \mathsurround\z@ \frozen@everymath{\mathgroup\macc@group\relax}%
  \macc@set@skewchar\relax
  \let\mathaccentV\macc@nested@a
  \macc@nested@a\relax111{#1}%
  \endgroup
}
\makeatother

\DeclareMathOperator{\diag}{diag}

\newcommand{\ifbcdot}[1]{\ifblank{#1}{\cdot}{#1}}

\DeclarePairedDelimiterX\abs[1]{\lvert}{\rvert}{\ifbcdot{#1}}
\DeclarePairedDelimiterX\parens[1]{(}{)}{\ifbcdot{#1}}
\DeclarePairedDelimiterX\brk[1]{[}{]}{\ifbcdot{#1}}
\DeclarePairedDelimiterX\braces[1]{\{}{\}}{\ifbcdot{#1}}
\DeclarePairedDelimiterX\angles[1]{\langle}{\rangle}{\ifblank{#1}{\cdot,\cdot}{#1}}
\DeclarePairedDelimiterX\ip[2]{\langle}{\rangle}{\ifbcdot{#1},\ifbcdot{#2}}
\DeclarePairedDelimiterX\norm[1]{\lVert}{\rVert}{\ifbcdot{#1}}
\DeclarePairedDelimiterX\ceil[1]{\lceil}{\rceil}{\ifbcdot{#1}}
\DeclarePairedDelimiterX\floor[1]{\lfloor}{\rfloor}{\ifbcdot{#1}}

\DeclareFontFamily{U}{matha}{\hyphenchar\font45}
\DeclareFontShape{U}{matha}{m}{n}{
      <5> <6> <7> <8> <9> <10> gen * matha
      <10.95> matha10 <12> <14.4> <17.28> <20.74> <24.88> matha12
      }{}
\DeclareSymbolFont{matha}{U}{matha}{m}{n}
\DeclareFontSubstitution{U}{matha}{m}{n}

\DeclareFontFamily{U}{mathx}{\hyphenchar\font45}
\DeclareFontShape{U}{mathx}{m}{n}{
      <5> <6> <7> <8> <9> <10>
      <10.95> <12> <14.4> <17.28> <20.74> <24.88>
      mathx10
      }{}
\DeclareSymbolFont{mathx}{U}{mathx}{m}{n}
\DeclareFontSubstitution{U}{mathx}{m}{n}

\DeclareMathDelimiter{\vvvert}{0}{matha}{"7E}{mathx}{"17}
\DeclarePairedDelimiterX\vertiii[1]{\vvvert}{\vvvert}{\ifbcdot{#1}}

\DeclarePairedDelimiterXPP\trace[1]{\operatorname{Tr}}{(}{)}{}{\ifbcdot{#1}} 
\DeclarePairedDelimiterXPP\col[1]{\operatorname{col}}{\{}{\}}{}{\ifbcdot{#1}} 
\DeclarePairedDelimiterXPP\row[1]{\operatorname{row}}{\{}{\}}{}{\ifbcdot{#1}} 
\DeclarePairedDelimiterXPP\erf[1]{\operatorname{erf}}{(}{)}{}{\ifbcdot{#1}}
\DeclarePairedDelimiterXPP\erfc[1]{\operatorname{erfc}}{(}{)}{}{\ifbcdot{#1}}
\DeclarePairedDelimiterXPP\KLD[2]{D}{(}{)}{}{\ifbcdot{#1}\, \delimsize\|\, \ifbcdot{#2}} 
\DeclarePairedDelimiterXPP\op[2]{\operatorname{#1}}{(}{)}{}{#2} 

\newcommand{\T}{^{\intercal}}

\DeclarePairedDelimiterXPP\indicate[1]{{\bf 1}}{\{}{\}}{}{\ifbcdot{#1}}

\providecommand\given{}

\DeclarePairedDelimiterX\Set[2]\{\}{%
\renewcommand\given{\SetSymbol[\delimsize]{#1}}
#2
}
\DeclarePairedDelimiterX\Setc[1]\{\}{%
\renewcommand\given{\SetSymbol{:}}
#1
}

\NewDocumentCommand\set{s o m}{%
	\IfBooleanTF#1%
	{\IfValueTF{#2}{\Set*{#2}{#3}}{\Setc*{#3}}}%
	{\IfValueTF{#2}{\Set{#2}{#3}}{\Setc{#3}}}%
}

\NewDocumentCommand{\evalat}{ s O{\big} m e{_^} }{%
\IfBooleanTF{#1}%
{\left. #3 \right|}{#3#2|}%
\IfValueT{#4}{_{#4}}%
\IfValueT{#5}{^{#5}}%
}

\providecommand\given{}
\DeclarePairedDelimiterXPP\cprob[1]{}(){}{
\renewcommand\given{\nonscript\,\delimsize\vert\allowbreak\nonscript\,\mathopen{}}%
#1%
}
\DeclarePairedDelimiterXPP\cexp[1]{}[]{}{
\renewcommand\given{\nonscript\,\delimsize\vert\allowbreak\nonscript\,\mathopen{}}%
#1%
}

\DeclareDocumentCommand \P { s e{_^} d() g } {%
	\mathbb{P}%
	\IfBooleanTF{#1}%
		{
			\IfValueT{#2}{_{#2}}%
			\IfValueT{#3}{^{#3}}%
			\IfValueTF{#5}{\cprob{#4 \given #5}}{\IfValueT{#4}{\cprob{#4}}}%
		}%
		{
			\IfValueT{#2}{_{#2}}%
			\IfValueT{#3}{^{#3}}%
			\IfValueTF{#5}{\cprob*{#4 \given #5}}{\IfValueT{#4}{\cprob*{#4}}}%
		}%
}

\DeclareDocumentCommand \E { s e{_^} o g } {%
	\mathbb{E}%
	\IfBooleanTF{#1}%
		{
			\IfValueT{#2}{_{#2}}%
			\IfValueT{#3}{^{#3}}%
			\IfValueTF{#5}{\cexp{#4 \given #5}}{\IfValueT{#4}{\cexp{#4}}}%
		}%
		{
			\IfValueT{#2}{_{#2}}%
			\IfValueT{#3}{^{#3}}%
			\IfValueTF{#5}{\cexp*{#4 \given #5}}{\IfValueT{#4}{\cexp*{#4}}}%
		}%
}

\ExplSyntaxOn
\NewDocumentCommand \dist {m o o} {%
\mathrm{#1}\left(%
	\IfValueT{#3}{%
		\tl_if_blank:nTF{ #3 }{\cdot\, \middle|\, }{#3\, \middle|\, }%
	}
	\IfValueT{#2}{#2}%
\right)%
}
\ExplSyntaxOff

\NewDocumentCommand {\cbrace} {t+ D[]{black} D(){\widthof{#5}} m m } {%
	\begingroup%
		\color{#2}
		\IfBooleanTF{#1}{%
			\overbrace{#4}^%
		}{
			\underbrace{#4}_%
		}%
		{\parbox[c]{#3}{\centering\footnotesize{#5}}}%
	\endgroup%
}

\let\oldforall\forall
\renewcommand{\forall}{\oldforall \, }

\let\oldexist\exists
\renewcommand{\exists}{\oldexist \, }

\graphicspath{{./Figures/}{./figures/}}
\DeclareDocumentCommand{\includeCroppedPdf}{ o O{./Figures/} m }{
	\IfFileExists{#2#3-crop.pdf}{}{%
		\immediate\write18{pdfcrop #2#3.pdf #2#3-crop.pdf}}%
	\includegraphics[#1]{#2#3-crop.pdf}
}

\makeatletter
\newcommand*{\addFileDependency}[1]{
  \typeout{(#1)}
  \@addtofilelist{#1}
  \IfFileExists{#1}{}{\typeout{No file #1.}}
}
\makeatother

\definecolor{gray90}{gray}{0.9}

\ifx\nohighlights\undefined

	\newcommand{\msout}[1]{\text{\color{green} \sout{\ensuremath{#1}}}}
	\newcommand{\del}[1]{{\color{green}\ifmmode \msout{#1}\else\sout{#1}\fi}}
\else

	\newcommand{\msout}[1]{#1}
	\newcommand{\del}[1]{#1}
\fi

\newcommand{\hhide}[1]{}

\newcommand{\txp}[2]{\texorpdfstring{#1}{#2}}

\ifx\diagnoselabel\undefined
	\relax
\else
	\makeatletter
	\def\@testdef #1#2#3{%
		\def\reserved@a{#3}\expandafter \ifx \csname #1@#2\endcsname
			\reserved@a  \else
			\typeout{^^Jlabel #2 changed:^^J%
				\meaning\reserved@a^^J%
				\expandafter\meaning\csname #1@#2\endcsname^^J}%
			\@tempswatrue \fi}
	\makeatother
\fi

\newcommand{\tb}[1]{\textbf{#1}}

\usepackage{ijcai23}

\usepackage{times}
\usepackage{soul}
\usepackage[hidelinks]{hyperref}
\usepackage[switch]{lineno}
\usepackage{amsthm}
\usepackage{float}

\newtheorem{theorem}{Theorem}
\theoremstyle{plain}

\theoremstyle{definition}
\newtheorem{definition}[theorem]{Definition}

\theoremstyle{remark}

\newcommand{\sd}[1]{\underline{\smash{#1}}}

\title{Graph Neural Convection-Diffusion with Heterophily }

\author{
    Author Name
    \affiliations
    Affiliation
    \emails
    email@example.com
}

\author{
Kai Zhao$^1$\textsuperscript{*}\and
Qiyu Kang$^1$\textsuperscript{*}\and
Yang Song$^2$\and
Rui She$^1$\and
Sijie Wang$^1$\And
Wee Peng Tay$^1$\\
\affiliations
$^1$Nanyang Technological University\\
$^2$C3 AI\\
\emails
\{kai.zhao,qiyu.kang\}@ntu.edu.sg, yang.song@c3.ai\\
\{rui.she,wang1679,wptay\}@ntu.edu.sg
}

\begin{document}

\maketitle

\begin{abstract}
    Graph neural networks (GNNs) have shown promising results across various graph learning tasks, but they often assume homophily, which can result in poor performance on heterophilic graphs. The connected nodes are likely to be from different classes or have dissimilar features on heterophilic graphs. In this paper, we propose a novel GNN that incorporates the principle of heterophily by modeling the flow of information on nodes using the convection-diffusion equation (CDE). This allows the CDE to take into account both the diffusion of information due to homophily and the ``convection'' of information due to heterophily. We conduct extensive experiments, which suggest that our framework can achieve competitive performance on node classification tasks for heterophilic graphs, compared to the state-of-the-art methods. The code is available at \url{https://github.com/zknus/Graph-Diffusion-CDE}.
\end{abstract}

\section{Introduction}
\footnotetext[1]{*Equal contribution.}
Graph Neural Networks (GNNs) have gained popularity in analyzing complex, structured data collected from various domains including social networks, citation networks, and molecular biology \cite{velickovic2017gat,JiLeeMen:labelgnn,KanZhaSon:hamgnn}. These datasets depict entities as nodes and relationships between entities as edges. Most pioneering GNN works like \cite{kipfgcn,velickovic2017gat,hamilton2017sage,xu2018gin,kipf2016vgae,Klibojste:appnp,LeeJiTay:simgat} assume the graph datasets have strong homophily, where linked nodes often belong to the same class or have similar features. These models fail to optimally utilize information in heterophilic graph datasets, where connected nodes are likely from distinct classes or possess dissimilar features. A study by \cite{sunhapyan:beyond2022} concludes that even simple models that do not take into account the graph structure, such as multilayer perceptrons (MLPs), can surpass several existing GNN models on heterophilic datasets. 

To tackle this challenge, several models like \cite{peiweicha:geomgcn2019,zhuyanhei:designs2020,chipenli:gprgnn2021,zhurosrao:graphheter,boxiashi:beyondlow2021,tanliliu:diverse2021,luahualu:revisit2022,crifraben:sheaf2022,chansudcam:simplified2022,sunhapyan:beyond2022,liyuche:find2022,maliusha:homo2021,wanjinrui:powerful,dushifu:gbkgnn2022} have been proposed. These works propose various techniques, such as aggregating higher-order neighborhoods, preserving high-frequency input signals, and adopting feature propagation methods, to improve the representation power of GNNs on heterophilic graph datasets.

Recently, works such as \cite{chamrowgor:grand2021,charoweyn:blend2021,SonKanWan:C22} have incorporated Neural Partial Differential Equations (PDEs) into GNNs, utilizing a variety of diffusion models for message passing on graphs. 
In this paper, our focus is on specific \emph{diffusion-based} graph PDE models to address graph heterophily, as opposed to more general graph PDEs like those presented in the study by \cite{ruscharow:graphcon2022}. The latter models nodes in a graph as coupled oscillators and establishes a second-order Ordinary Differential Equation (ODE) for updating node features.
In \cite{chamrowgor:grand2021}, the authors model information propagation as a diffusion process of a substance from regions of higher to lower concentration, such as heat diffusion from a hot object to a cold surface. However, due to the nature of heat diffusion, the features of neighboring nodes tend to become increasingly smooth. In heterophilic graphs, where connected nodes are likely to have dissimilar features, this method of message passing may not be appropriate. 
The paper \cite{crifraben:sheaf2022} proposes more general sheaf diffusion operators to control the diffusion process and maintain the non-smoothness in heterophily graphs, resulting in improved node classification performance. 
Furthermore, the study \cite{zhu2022does} reveals a strong connection between heterophily and the robustness of GNNs. Recently, research has drawn inspiration from works like \cite{KanSonDinTay:stableode,SonKanTay:errorcorr} to explore how specially designed diffusion processes could enhance the robustness of GNNs against adversarial attacks. This has been demonstrated in the context of both classification \cite{SonKanWan:C22} and localization tasks \cite{WanKanShe:robustloc}. Further research on diffusion models for heterophilic graphs is of particular interest. 

 In this paper, we propose a novel GNN inspired by the convection-diffusion process. Our proposed model explicitly incorporates the principle of heterophily in its design, which differentiates it from existing graph neural diffusion models. In a convection-diffusion heat transfer process in a fluid or gas, heat is not only transferred through diffusion due to heat concentration gradients but also through the movement of the fluid or gas itself. The velocity term in convection can be thought of as the movement of heat energy from one place to another due to the physical movement of a fluid or gas. The curvature-preserving diffusion models such as Beltrami and mean-curvature \cite{charoweyn:blend2021,SonKanWan:C22} are able to slow down the diffusion at the nodes where the feature difference between neighboring nodes is large, thereby preserving the non-smoothness and network robustness. Our model generalizes these curvature-preserving models by introducing a convection term in the diffusion equation that explicitly controls the propagation velocity at each node.

 \tb{Main contributions.} In this paper, our objective is to develop a general diffusion-based graph PDE framework that is suitable for heterophilic graph datasets. Our main contributions are summarized as follows: 
 \begin{enumerate}
    \item Based on the convection-diffusion equation (CDE), we propose a graph neural convection-diffusion model that includes both a diffusion and convection term. The diffusion term aggregates information from homophilic neighbors, while the convection term allows controlled information propagation from heterophilic neighbors.
    \item To manage the convection term in our CDE model, we introduce a learnable mechanism that adjusts the propagation rate at every node, allowing for better handling of heterophily. 
    \item We conduct extensive experiments on benchmark heterophilic datasets and compare with state-of-the-art baselines. Our experiments demonstrate that our model is competitive compared to the baselines, especially on large and more heterophilic datasets.
 \end{enumerate}

\section{Related Work}
In what follows, we briefly review the GNNs that are proposed to handle heterophilic graphs and recent advances in graph neural diffusion models.

\subsection{Graph Neural Networks with Heterophily} \label{subsect:GNN_heter}
Several works aim to adapt normal GNNs to heterophilic networks. H2GCN \cite{zhuyanhei:designs2020} proposes three effective designs to boost the performance of GNNs on heterophilic graphs, including the ego- and neighbor-embedding separation, higher-order neighborhood aggregation, and the combination of intermediate representations. Geom-GCN \cite{peiweicha:geomgcn2019} maps the graph to a suitable latent space where it can aggregate immediate neighborhoods and distant nodes to preserve the topology patterns of the graph. GPR-GNN \cite{chipenli:gprgnn2021} assigns each step of feature propagation with a learnable weight, which can be negative or positive, to mitigate graph heterophily and over-smoothing issues. CPGNN \cite{zhurosrao:graphheter} incorporates a compatibility matrix into GNNs to learn the likelihood of connections between nodes in different classes, which allows a GNN to capture both heterophily and homophily patterns in the graph. FAGCN \cite{boxiashi:beyondlow2021} integrates the low-frequency  signals, high-frequency signals, and raw features adaptively through a self-gating mechanism to enhance the expressive power of GNNs on disassortative networks. ACM-GCN \cite{luahualu:revisit2022}  adaptively learns the local and node-wise information through aggregation, diversification, and identity channels. In this way, it can retain the low-frequency and high-frequency components of the input signal. Instead of using fixed feature propagation step in SGC \cite{wuaouzha:sgc2019}, ASGC \cite{chansudcam:simplified2022} applies a learned polynomial of the normalized adjacency matrix with the input feature to fit a different filter for each feature. Our work is different from these previous works as we mainly focus on the use of graph PDEs to model the information propagation in heterophilic graphs.

\subsection{Graph Neural Diffusion}
In this subsection, we provide a brief overview of various graph neural diffusion models that have been proposed recently.  In \cite{chamrowgor:grand2021}, the authors modeled information propagation as a diffusion process of a substance from regions of higher to lower concentration. The Beltrami diffusion model was utilized in \cite{charoweyn:blend2021,SonKanWan:C22} to enhance rewiring and improve the robustness of the graph. The above works do not specifically target the problem of graph heterophily. 

To address diffusion in heterophilic graphs, the paper \cite{crifraben:sheaf2022} introduces general sheaf diffusion operators to control the diffusion process and maintain non-smoothness in heterophilic graphs, leading to improved node classification performance. Inspired by the particle reaction-diffusion process, the ACMP\cite{wang2023acmp} models the heterophilic graph by employing both repulsive and attractive force interactions as dual flow directions between nodes. 
 Our approach differs from the aforementioned diffusion models in that we incorporate a convection term that is able to adapt to the complex connections in heterophilic graphs.

\section{Preliminaries}\label{sec:Preliminaries}
In this section, we first introduce the convection-diffusion equation (CDE). We then review graph neural diffusion formulations that are used in our graph CDE model and will be later used in experiments on graph datasets. 

\subsection{Convection-Diffusion Equation}
In \cite{chamrowgor:grand2021}, the authors unified many popular GNN architectures into a single mathematical framework that is derived from the following (heat) diffusion equation. Let $x(u, t)$ denote a scalar-valued function on $\Omega \times [0, \infty)$, e.g., $x(u, t)$ can be the temperature at point $u$ on the space $\Omega$ at time $t$. The heat diffusion \cite{grigoryan2009heat} process is formulated as 
\begin{align}
\frac{\partial x}{\partial t}= \operatorname{div} ( D \nabla x),\ t>0,
\label{eq:heat_cauchy}
\end{align}
with some initial boundary condition that specifies the temperature distribution at $t=0$, and $\operatorname{div}$ and $\nabla$ are the divergence and concentration gradient operators, respectively. 
Here $D$ is the thermal diffusivity for heat diffusion, which may vary at different positions in an inhomogeneous material and leads to position-dependent diffusion directions and speeds. In the GNN models \cite{SonKanWan:C22,chamrowgor:grand2021}, the authors design different formulations $D$ to model the information propagation between graph nodes that may depend on the node and its neighbors.

The CDE, on the other hand, describes physical phenomena such as heat transfer within a physical system, resulting from the combination of two processes: diffusion and convection. The CDE is given by
\begin{align}
\frac{\partial x}{\partial t}= \operatorname{div} ( D \nabla x)  -  \operatorname{div} ( \bv x)
\label{eq:cde}
\end{align}
where the first term represents the heat diffusion, same as in \cref{eq:heat_cauchy}, and the second term $\operatorname{div} (\bv x)$ in the equation represents the convection process. The velocity field $\bv$ is a function of both time and space and describes how the quantity is moving. In oceanography, the convection-diffusion equation is used to model the transfer of heat in the ocean. The variable $x$ would represent the concentration of heat in the ocean, and the velocity field $\bv$ would represent the ocean currents and water flow as a function of both time and location. 

\subsection{Common Graph Neural Diffusion Models}\label{sec.comm_graph_diff}
Suppose $\calG = (\calV, \calE, w)$ is a graph, where $\calV = \{1, \dots, N\}$ is the set of $N$ nodes, $\calE \subseteq \mathcal{V} \times \mathcal{V}$ is the set of edges, and $w: \calE \rightarrow \Real^+$ the edge weight function. The features for the vertices at time $t$ are represented by $\bX(t) \in \Real^{|\calV|\times r}$ where $r$ is the node feature dimension. The $i$-th row of $\bX(t)$, denoted as $\bx\T_i(t)$, is the feature vector for vertex $i$ at time $t$.

In the literature, several graph diffusion models \cite{chamrowgor:grand2021,SonKanWan:C22,crifraben:sheaf2022} have been proposed, which are mostly derived from the heat equation \cref{eq:heat_cauchy} with different diffusivities $D$ learned from the graph datasets. For example, inspired from the heat diffusion equation \cref{eq:heat_cauchy}, GRAND \cite{chamrowgor:grand2021} defines the following graph version of the gradient $\nabla$ and divergence $\operatorname{div}$ operators as
\begin{align}
    (\nabla \bX(t))_{(i,j)}= \bx_j(t)-\bx_i(t), \ \forall (i,j)\in \calE
\end{align}
for each edge $(i,j)\in \calE$ in the graph, and
\begin{align}
    (\operatorname{div}(\mathscr{X}))_i=\sum_{j:(i,j)\in \calE} \mathscr{X}_{i j},
\end{align}
where $\mathscr{X}=\{\mathscr{X}_{i j}\}_{(i,j)\in \calE}$ is the set of all features associated with edges and $\mathscr{X}_{i j}\in \Real^r$ is the feature associated with edge $(i,j)$. Furthermore, for graph learning, they implement the following dynamical system:
\begin{align}
\frac{{\partial} \bX(t)}{{\partial} t} 
&= \operatorname{div}(D(\bX(t), t) \odot \nabla \bX(t))\nn
& = (\mathbf{A}(\bX(t))-\mathbf{I}) \bX(t)  \label{eq.GRAND}
\end{align} 
where the initial condition is given by $\bX(0)=\left(\bx^{\T}_0(0), \cdots,\bx^{\T}_{|\calV|}(0)\right)^{\T}$, $\odot$ is the element-wise product and the diffusivity $D$ is a $|\calE|\times |\calE|$ diagonal matrix with elements $\diag(a(\bx_i(t), \bx_j(t), t))$. Here, $a(\cdot)$ is a similarity function for vertex pairs. The diffusion equation can therefore be reformed as \cref{eq.GRAND} with matrix $\mathbf{A}(\bX(t))= \left(a\left(\bx_i(t),  \bx_j(t)\right)\right)$, which is a learnable attention matrix used to describe the structure of the graph, and $\mathbf{I}$ being an identity matrix.

We refer the readers to the supplementary material for more details about other diffusion variants like the graph Beltrami flow \cite{SonKanWan:C22}, and graph sheaf diffusion \cite{crifraben:sheaf2022}.

\subsection{Graph Heterophily}

In this section, we introduce the graph homophily measure, which is used in the experiments in \cref{sec.exp} to quantify the homophily level of a graph dataset.
\begin{definition}[Edge homophily ratio \cite{zhuyanhei:designs2020}] \label{dfn:homophily-ratio}
    The edge homophily ratio is defined as: 
    \begin{equation}\label{eq:edge-homophily}
    h_{\mathrm{edge}} = \frac{|\set*{(u,v) \given (u,v) \in \calE \wedge y_u = y_v}|}{|\calE|}.
    \end{equation}
    This ratio is the proportion of edges in a graph that connects nodes with the same class label (i.e., intra-class edges). A smaller $h_{\mathrm{edge}}$ implies stronger heterophily. 
\end{definition}

The edge homophily ratio is sensitive to the number of classes and the balance of classes in a graph. To fairly compare the homophily ratio across different datasets with different numbers of classes and class balance, the paper \cite{plakuzbab:adjusted2022} proposes the adjusted homophily ratio as follows.

\begin{definition}[Adjusted homophily ratio \cite{plakuzbab:adjusted2022}] \label{dfn:adjustedhomophily-ratio}
   
    Suppose there are $C$ classes. The adjusted homophily ratio is given by
    \begin{align}\label{eq:adjusted-homophily}
         h_{\mathrm{adj}} = \frac{h_{\mathrm{edge}} - \sum_{k=1}^C D_k^2/(2 |\calE|)^2}{1 - \sum_{k=1}^C D_k^2/(2 |\calE|)^2} 
    \end{align}
    where $D_k:=\sum_{i: y_i = k} d(i)$, $y_i$ is the label of node $i$, and $d(i)$ is the degree of node $i$.
\end{definition}

\section{Graph Neural Convection-Diffusion}
We now consider graph neural convection-diffusion by making use of concepts from \cref{sec:Preliminaries}. 
 In homophilic graphs, the edges between nodes tend to be between similar nodes, analogous to homogeneous atoms or molecules in a solid material. In such materials, these connections are relatively stable. The movement of atoms or molecules is relatively limited by the strength of the connections between them, and the heat transfer is mainly described using \cref{eq:heat_cauchy}. Analogously, the information aggregation in homophilic graphs can be well-described by the graph heat diffusion equation \cite{chamberlain2021grand}.
 
 In contrast, heterophilic graphs have connections between dissimilar nodes, analogous to having different types of particles in a gas or fluid. These connections are less stable, and the movement of particles is relatively unrestricted. This makes heterophilic graphs well-suited to be regarded as soft materials, where the movement of particles is not limited by the strength of their connections.
In the CDE \cref{eq:cde}, the velocity term (convection) is used to model the movement of the fluid or gas itself. 
For information propagation in heterophilic graphs, the convection term can help us to assign different information propagation velocities to each node due to the different connections.

The graph analogy of \cref{eq:cde} can be formulated as
\begin{align}
   \frac{\partial}{\partial t} \bX (t)=  \operatorname{div}(D(\bX(t), t) \odot \nabla \bX(t)) + \operatorname{div}(\bV(t) \circ \bX(t))\label{eq:graph_cde}
\end{align}
where the first diffusion term $\operatorname{div}(D(\bX(t), t) \odot \nabla \bX(t)) $ could be given by \cref{eq.GRAND} or other variants in \cite{SonKanWan:C22,crifraben:sheaf2022}, $\bV_{ij}(t)\in \Real^r$ is the velocity vector associated with each edge $(i,j)$ at time $t$, $\bV(t) = \{\bV_{ij}(t) \}_{(i,j)\in\calE}$, and 
\begin{align}\label{eq.conv_term}
    (\operatorname{div}(\bV(t) \circ \bX(t)))_i \coloneqq \sum_{j:(i, j) \in \mathcal{E}} \bV_{ij}(t) \odot  \bx_j(t)
\end{align}
for each node $i \in \calV$.

\subsection{\txp{$\bV_{ij}$}{Vij} for Heterophily}\label{sec.vij}
We propose the following simple yet effective velocity formulation:
\begin{align}\label{eq.rgdadfa}
    \bV_{ij}=  \sigma\left(W(\bx_j(t)-\bx_i(t))\right),
\end{align}

where $W$ is a learnable matrix and $\sigma$ denotes an activation function.
In \cref{eq.rgdadfa}, the velocity $\bV_{ij}$ determines the direction of preferred information transport from node $j$ to its neighboring node $i$. 
The velocity is determined by the difference $\bx_j(t)-\bx_i(t)$ of the feature representations of the neighboring nodes.  Note that the velocity is not always in the same direction as the difference $\bx_j(t)-\bx_i(t)$ due to the learnable $W$ and activation function.  
The use of the velocity term in the dot product with the state of the neighboring node in \cref{eq.conv_term} allows for a more flexible and adaptive information flow for heterophilic graphs. This is because it takes into account the dissimilarity of neighboring nodes, which is a unique characteristic of heterophilic graphs. In contrast, traditional neural PDEs \cite{chamrowgor:grand2021,SonKanWan:C22} utilize only a single diffusion process, which does not consider the dissimilarity of neighboring nodes. The addition of the convection term in the CDE allows for a more accurate representation of the information flow in heterophilic graphs. For a more comprehensive understanding of our model, we direct readers to the supplementary material.

\subsection{Model Details}
We present our algorithm in \cref{alg:geognn}. Our approach is based the graph neural PDE defined in \cref{eq:graph_cde}. This graph PDE is distinct from the diffusion-based models presented in previous works such as \cite{chamrowgor:grand2021,SonKanWan:C22,crifraben:sheaf2022}. To solve the neural PDE, we follow the convention in these previous works and use the provided solver from \cite{chen2018neural}. Our main contribution is the introduction of the second term in \cref{eq:graph_cde}, which utilizes a simple yet effective velocity formulation \cref{eq.rgdadfa}). The first term in \cref{eq:graph_cde} has various options as discussed in \cite{SonKanWan:C22,chamberlain2021grand}. 
\begin{algorithm}[!htb]
\caption{Neural CDE Inference}\label{alg:geognn}
\begin{algorithmic}[1]
\Require $\calG = (\calV, \calE)$, integration time $T$.
\Ensure  Graph task results
\State Compress the sparse raw input node feature using an MLP and set them as the initial conditions $\{\bx_i(0)\}_{i=1}^{|\calV|}$.
\State Solve the neural PDE \cref{eq:graph_cde} using explicit or implicit solvers from\cite{chen2018neural}.
\State Return $\{\bx_i(T)\}_{i=1}^{|\calV|}$ which is the solution of \cref{eq:graph_cde}  at time $T$.
\State Perform further graph tasks, e.g., node classification. 
\end{algorithmic}
\end{algorithm}

\section{Experiments}
\label{sec.exp}

\begin{table*}[!ht]
\centering
\resizebox{0.9\textwidth}{!}{\begin{tabular}{ccccccc}
\toprule
 Dataset & Nodes & Edges & Classes & Node Features & Edge homophily $h_{\mathrm{edge}}$  & Adjusted homophily $ h_{\mathrm{adj}}$  \\
 \midrule
 Roman-empire &  22662     &  32927     &      18   &  300 & 0.05 & -0.05      \\
 Wiki-cooc &  10000     &  2243042     &      5   &  100 & 0.34 & -0.03     \\
 Minesweeper &  10000     &  39402     &     2   &  7  & 0.68 & 0.01    \\
 Questions &  48921     &  153540     &     2   &  301 & 0.84 & 0.02     \\
 Workers &  11758     &  519000     &     2   &  10  & 0.59 & 0.09    \\
 Amaon-ratings &  24492     &  93050     &     5   &  300  & 0.38 & 0.14    \\
 \hline 
Texas &  183     &  295     &     5   &  1703 & 0.11  & 0.04    \\
 Cornel &  183     &  280    &     5  &  1703 & 0.30 & 0.04   \\
 Wisconsin &  251    &  466     &     5   &  1703 & 0.21 & 0.07   \\


 \bottomrule
\end{tabular}}
\caption{Bechmark's statistics}
\label{tab:bechmark}
\end{table*}

\begin{table*}[!htp]\small
\centering
 \resizebox{1\textwidth}{!}{\begin{tabular}{c|cccccc|ccc}
\toprule
Method   & Roman-empire      & Wiki-cooc           & Minesweeper      & Questions             & Workers            & Amazon-ratings          & Texas & Cornell & Wisconsin   \\
$h_{\mathrm{adj}}$ &-0.05        &-0.03              & 0.01             & 0.02               & 0.09                  & 0.14                  &0.04 &0.04 & 0.07  \\
\midrule
ResNet  &  65.71$\pm$0.44     &  89.36$\pm$0.71     &  50.95$\pm$1.12   & 70.10$\pm$0.75      &  73.08$\pm$1.28     &  45.70$\pm$0.69     &  80.81$\pm$4.75  &  81.89$\pm$6.40  &  85.29$\pm$3.31  \\

\midrule
GCN    &  70.51$\pm$0.75        &  91.01$\pm$0.69    &  89.47$\pm$0.69   & 75.97$\pm$1.23       &\sd{83.14$\pm$1.11}  & 47.77$\pm$0.69    &  55.14$\pm$5.16  &  60.54$\pm$5.30  &  51.76$\pm$3.06    \\

GraphSAGE  &\sd{85.80$\pm$0.69} &  93.60$\pm$0.31     &  93.53$\pm$0.50   & 76.52$\pm$0.93       &  82.34$\pm$0.19    & \tb{52.77$\pm$0.54}  &  82.43$\pm$6.14  &  75.95$\pm$5.01  &  81.18$\pm$5.56 \\
GAT     &  81.02$\pm$0.46       &  92.44$\pm$0.80    &  92.10$\pm$0.67    & \tb{77.42$\pm$1.23}   &\tb{83.98$\pm$0.57}  &\sd{47.95$\pm$0.53}  &  52.16$\pm$6.63  &  61.89$\pm$5.05  &  49.41$\pm$4.09   \\

\midrule
H2GCN   &  68.09$\pm$0.29       & 89.24$\pm$0.32    &  89.95$\pm$0.38     &66.66$\pm$1.84           & 81.76$\pm$0.68     & 41.36$\pm$0.47   & 84.86$\pm$7.23  &  82.70$\pm$5.28  &  87.65$\pm$4.98      \\
CPGNN   &  63.78$\pm$0.50       & 84.84$\pm$0.66    &  71.27$\pm$1.14     & 67.09$\pm$2.63          & 72.44$\pm$0.80     & 44.36$\pm$0.35   & 75.68$\pm$5.12  &  70.27$\pm$5.12  &  76.47$\pm$6.16      \\

GPR-GNN  &  73.37$\pm$0.68      & 91.90$\pm$0.78     &  81.79$\pm$0.98    & 73.41$\pm$1.24         &  70.59$\pm$1.15    &  43.90$\pm$0.48   &  81.35$\pm$5.32  &  78.11$\pm$6.55  &  82.55$\pm$6.23     \\

GloGNN  &  63.85$\pm$0.49       &  88.49$\pm$0.45     &  62.53$\pm$1.34   &  67.15$\pm$1.92        &  73.90$\pm$0.95    &  37.28$\pm$0.66    &  84.32$\pm$4.15  &  83.51$\pm$4.26  &  87.06$\pm$3.53    \\
FAGCN  &  70.53$\pm$0.99          &  91.88$\pm$0.37   &  89.69$\pm$0.60     & \sd{77.04$\pm$1.56}    &  81.87$\pm$0.94    &  46.32$\pm$2.50  &  82.43$\pm$6.89  &  79.19$\pm$9.79  &  82.94$\pm$7.95      \\
GBK-GNN   &  75.87$\pm$0.43       &  \sd{97.81$\pm$0.32}    &  83.56$\pm$0.84    & 72.98$\pm$1.05        &  78.06$\pm$0.91   &  43.47$\pm$0.51   &  81.08$\pm$4.88  &  74.27$\pm$2.18  &  84.21$\pm$4.33      \\
ACM-GCN   &  68.35$\pm$1.95      &  87.48$\pm$1.06    &  90.47$\pm$0.57     & OOM                  &  78.25$\pm$0.78   &  38.51$\pm$3.38     &  \tb{87.84$\pm$4.40}  &  85.14$\pm$6.07  &  \sd{88.43$\pm$3.22}    \\

\midrule
GRAND  &  71.60$\pm$0.58         & 92.03$\pm$0.46     & 76.67$\pm$0.98        & 70.67$\pm$1.28   &  75.33$\pm$0.84    &  45.05$\pm$0.65     &  80.27$\pm$4.84  &  82.97$\pm$6.63  &  83.73$\pm$3.51     \\

GraphBel &  69.47$\pm$0.37     &  90.30$\pm$0.50     &  76.51$\pm$1.03     & 70.79$\pm$0.99         &  73.02$\pm$0.92  &43.63$\pm$0.42     &  80.27$\pm$3.64  &  83.51$\pm$6.45  &  85.29$\pm$4.82    \\

Diag-NSD  &  77.50$\pm$0.67     &  92.06$\pm$0.40    &  89.59$\pm$0.61     & 69.25$\pm$1.15         &  79.81$\pm$0.99   &  37.96$\pm$0.20    &  85.67$\pm$6.95  &  \tb{86.49$\pm$7.35}  &  \tb{88.63$\pm$2.75}      \\
ACMP&  71.27$\pm$0.59 & 92.68$\pm$0.37  & 76.15$\pm$1.12  & 71.18$\pm$1.03 & 75.03$\pm$0.92  &44.76$\pm$0.52  & 86.20$\pm$3.0 & 85.40$\pm$7.0 &  86.10$\pm$4.0 \\
\midrule
CDE-GRAND  &  \tb{91.64$\pm$0.28}  &\tb{97.99$\pm$0.38} &\tb{95.50$\pm$5.23} & 75.17$\pm$0.99       &  80.70$\pm$1.04  &47.63$\pm$0.43    & 86.22$\pm$3.30  &  \sd{86.22$\pm$5.05}  &  87.45$\pm$4.40  \\
CDE-GraphBel  &  85.39$\pm$0.46    &97.79$\pm$0.40     &\sd{93.98$\pm$0.57}  & 72.11$\pm$1.31       &  81.30$\pm$0.43  &45.22$\pm$0.60     & \sd{87.57$\pm$3.24}  &  85.14$\pm$5.95  &  87.84$\pm$4.87  \\

\bottomrule
\end{tabular}}
\caption{Node classification results(\%). The best and the second-best result for each criterion are highlighted in \tb{bold} and \sd{underlined} respectively. The results are accuracy on Wiki-cooc, Roman-empire, Amaon-ratings,  Texas, Cornell and Wisconsin datasets and ROC-AUC score on Minesweeper, Workers, and Questions datasets. OOM refers to out-of-memory on NVIDIA RTX A5000 GPU. }
\label{tab:noderesults}
\end{table*}

\subsection{Real-World Datasets} \label{sec:datasets}
The paper \cite{peiweicha:geomgcn2019} evaluates the performance of their model on six heterophilic graph datasets: Squirrel, Chameleon, Actor, Texas, Cornell, and Wisconsin. These are widely-used heterophilic benchmarks, and we also use them to evaluate different models' node classification performance. 

However, a recent paper \cite{Data_paper} has pointed out a train-test data leakage problem in the Squirrel and Chameleon datasets collected by \cite{rozcarrik:multi2021}, leading to questions on model performance on these two datasets. The datasets Cornell, Texas, and Wisconsin \footnote{Available in http://www.cs.cmu.edu/afs/cs.cmu.edu/project/theo-11/www/wwkb} do not have this data leakage issue but are relatively small and have significantly imbalanced classes, which can result in overfitting for diffusion models \cite{crifraben:sheaf2022}. We report the node classification results based on these three datasets in \cref{tab:noderesults}, where we use a fixed data splitting method as introduced in \cite{peiweicha:geomgcn2019}. The experiments where we apply random data splitting are provided in the Appendix, where we also include experiments using Squirrel, Chameleon, and Actor datasets. 

To fairly evaluate the model performance on heterophilic graph datasets, we additionally include six new heterophilic datasets proposed in \cite{Data_paper}.
The proposed datasets come from different domains, such as English Wikipedia, the Amazon product co-purchasing network, and a question-answering platform. All of them have low homophily scores as verified in \cite{Data_paper} and diverse structural properties. 
\cite{Data_paper} evaluates a wide range of GNNs, both standard, and heterophily-specific, on these datasets and argues that the progress in graph learning under heterophily is limited since the standard baselines usually outperform heterophily-specific models on the datasets.
For these heterophilic datasets, we follow the data splitting in \cite{Data_paper}, which is 50\%, 25\%, and 25\% for training, validation, and testing. The data statistics are shown in \cref{tab:bechmark}. 
We use the ROC-AUC score as the evaluation metric on Minesweeper, Workers, and Questions datasets because they have binary classes. We include the synthetic regular graphs using the algorithm described in \cite{luahualu:revisit2022} to control dataset heterophily for further demonstration.


\subsection{Implementation and Baselines}
We evaluate our neural CDE model on the node classification task against a variety of baseline models, including a graph-agnostic model, ResNet \cite{he2016resnet}, as well as standard GNN models such as GCN \cite{kipfgcn}, GAT \cite{velickovic2017gat}, and GraphSAGE \cite{hamilton2017sage}. Additionally, we compare our model to GNNs specifically designed for heterophilic graphs, including H2GCN \cite{zhuyanhei:designs2020}, CPGNN \cite{zhurosrao:graphheter}, GPR-GNN \cite{chipenli:gprgnn2021}, GloGNN \cite{liyuche:find2022}, FAGCN \cite{boxiashi:beyondlow2021}, GBK-GNN \cite{dushifu:gbkgnn2022}, and ACM-GCN\cite{luahualu:revisit2022}. We also include diffusion-based graph PDE models such as GRAND \cite{chamberlain2021grand}, GraphBel \cite{SonKanWan:C22}, Diag-NSD \cite{crifraben:sheaf2022} and ACMP \cite{wang2023acmp} as baselines.

We have implemented our CDE models using the algorithm outlined in \cref{alg:geognn}. In this model, the first diffusion term in \cref{eq:graph_cde} is specified by the graph diffusion formulation presented in previous works such as GRAND and GraphBel \cite{chamrowgor:grand2021,SonKanWan:C22}. We provide detailed descriptions of the specific diffusion formulations used in our implementation in \cref{sec.variousdiff_f}. Specifically, we refer to our CDE model that utilizes the diffusion formulation in \cref{eq:bahdanau_attention} as CDE-GRAND-GAT. For simplicity, we may use the abbreviation CDE-GRAND to refer to it in the rest of the paper. CDE-GraphBel refers to the CDE model with GraphBel being the first term.
  For all these datasets, we use the Adam optimizer \cite{kingma2014adam} with a learning rate of 0.01 and weight decay of 0.001. We also apply a dropout rate of 0.2 to prevent overfitting issues. We solve the neural PDE through the methods in \cite{chen2018neural}. The integration time and step size are adjusted according to the performance of validation data. For the Texas, Cornell, and Wisconsin datasets, which have much smaller sizes, we conduct the hyperparameter search on each dataset. We report the best hyperparameters of each dataset in the supplementary material.

\subsection{Node Classification Results on Real-World Datasets}
From \cref{tab:noderesults}, we observe that if we augment diffusion models like GRAND and GraphBel with our proposed CDE convection term \cref{eq.conv_term}, we can significantly improve the diffusion model performance on the heterophilic graph datasets.
For example, on the Roman-empire dataset, CDE-GRAND improves GRAND's performance from 71.60$\pm$0.58\% to 91.64$\pm$0.28\%. Similar improvements can be observed on \emph{all of the nine heterophilic graph datasets}, demonstrating the effectiveness of our convection term in addressing the unique challenges of heterophilic graphs.
This is because our velocity takes into account the dissimilarity of neighboring nodes in heterophilic graphs. 
For information propagation in heterophilic graphs, the convection term that includes the dot product between the neighboring node and the velocity helps us to assign flexible and adaptive information propagation flows to each node due to the different connections. 
We refer the readers to \cref{sec.variousdiff_f} for further demonstration.

Compared with all the baselines on heterophilic graph datasets, CDE-GRAND achieves the best performance on Wiki-cooc, Roman-empire, and Minesweeper. 
From \cref{tab:bechmark}, we observe that these datasets have the lowest adjusted homophily $h_{\mathrm{adj}}$, indicating they are highly heterophilic. Our results suggest that CDE diffusion is particularly effective on more heterophilic datasets.
On the Roman-empire dataset, CDE-GRAND outperforms the second-best model, GraphSAGE, by a significant margin of 5.84\%. Notably, GraphSAGE is a standard GNN that is not specifically designed for heterophilic graphs, yet it performs better than all heterophily-specific baselines, highlighting the current limitations in graph learning under heterophily \cite{Data_paper}. 
However, our CDE-GRAND model successfully surpasses GraphSAGE on this heterophilic dataset. 
A similar trend can be observed in the Minesweeper dataset where GraphSAGE achieves 93.53$\pm$0.50\% and outperforms all the heterophily-specific baselines. Our CDE-GRAND model, however, can surpass GraphSAGE by a significant margin of 4.14\%.
On all other heterophilic graph datasets, CDE-GRAND performs comparably to state-of-the-art baselines, with the exception of Amazon-ratings, where all heterophily-specific models perform worse than GraphSAGE. This may be because the Amazon-ratings dataset has less heterophilic structure, as indicated by the largest adjusted homophily $ h_{\mathrm{adj}}$ score presented in \cref{tab:bechmark}. 
As observed in \cite{Data_paper} and \cref{tab:noderesults}, the modified GAT with residual connections outperforms all other heterophily-specific GNN models on the Questions, Workers, and Amazon-ratings datasets, hinting at unique geometries in these datasets that other models fail to capture effectively. 

\begin{figure}[!ht] 
\centering
\includegraphics[width=0.84\linewidth]{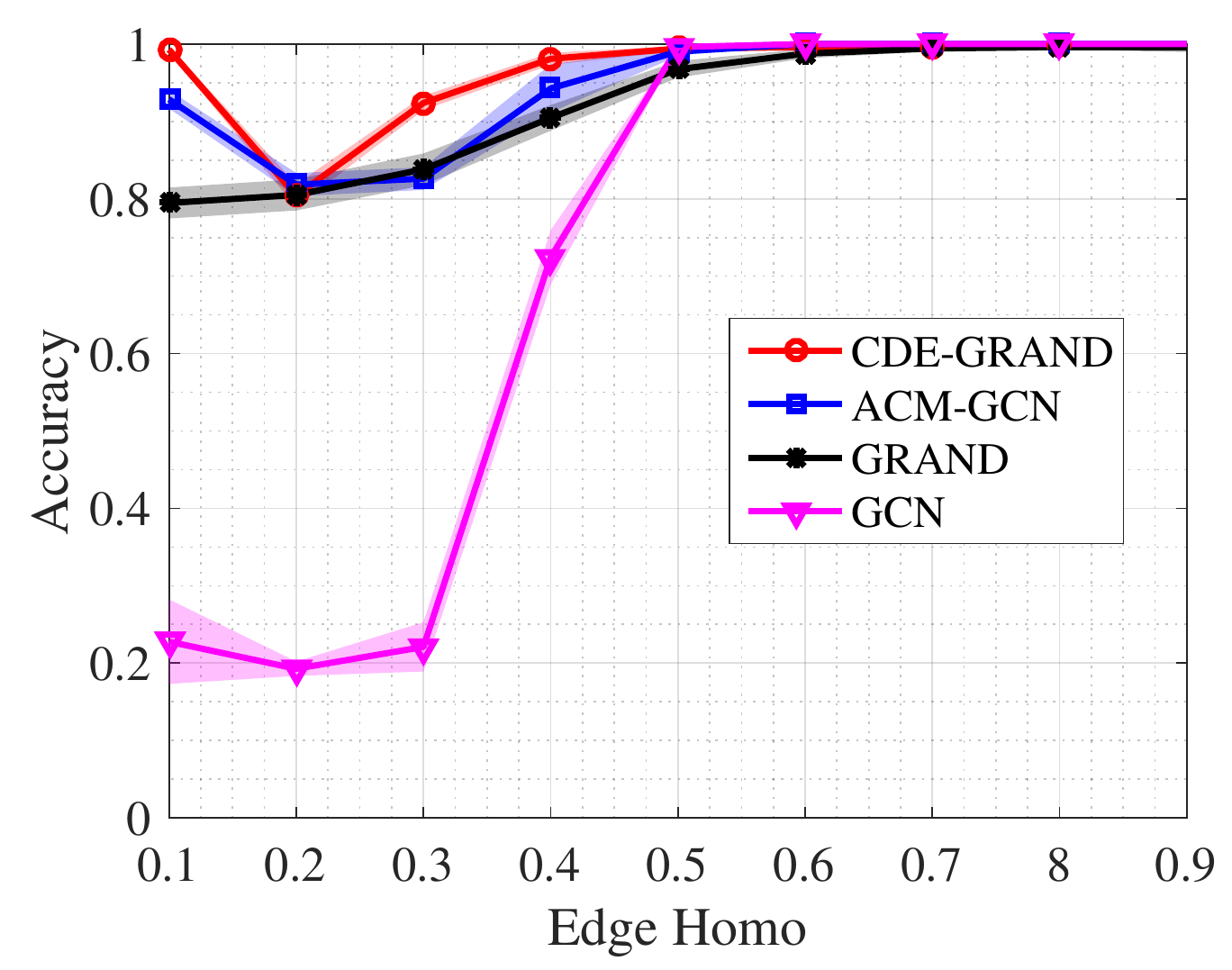}
\caption{Accuracy versus edge homophily level. A node classification task based on the Cora dataset is used to compare different methods including CDE-GRAND, ACM-GCN, GRAND, and GCN.}
\label{fig:edge_homo}
\end{figure}
\subsection{Node Classification Results on Synthetic Datasets}
\label{subs:Syn}
We generate synthetic regular graphs using the algorithm described in \cite{luahualu:revisit2022}. The edge homophily levels range from 0.1 to 0.9, with 10 graphs generated for each level. Each graph contains 5 classes, with 400 nodes in each class. For each node, we randomly generate 10 edges to connect to nodes within the same class and [$\frac{10}{H_\text{edge}(\mathcal{G})} -10$] edges to nodes in other classes. The features of each node are sampled from the corresponding class of the base dataset, which is the Cora dataset in \cref{fig:edge_homo}. The nodes are then randomly split into train, validation, and test sets at a ratio of 60\%:20\%:20\%. The node classification results are displayed in \cref{fig:edge_homo}. 

From \cref{fig:edge_homo}, we observe that CDE-GRAND outperforms other methods on classification accuracy. Especially when the edge homophily level is less than 0.5, the CDE-GRAND has advantages obviously. This indicates that our method is more adaptive to the heterophily of graphs than the other methods including ACM-GCN, GRAND, and GCN.

\begin{table}[!htb]\small
    \centering
    \begin{tabular}{cccc}
    \toprule
        Time &  ODE Solver & Roman-empire & Minesweeper  \\
        \midrule 
        
         \multirow{2}{*}{1.0} & Euler Solver & 87.26$\pm$0.46 & 87.13$\pm$1.36  \\ 
                             & RK4 Solver & 91.55$\pm$0.66 & 93.05$\pm$0.48 \\ 
         \midrule
         \multirow{2}{*}{2.0} & Euler Solver & 91.55$\pm$0.42 & 90.46$\pm$0.66  \\ 
                             & RK4 Solver & 91.10$\pm$1.16 & 94.64$\pm$2.32\\ 
         \midrule
         \multirow{2}{*}{3.0} & Euler Solver & 91.64$\pm$0.28 & 92.00$\pm$0.60\\
                             & RK4 Solver & 90.82$\pm$2.18 & 94.35$\pm$3.62   \\ 
         \midrule
        \multirow{2}{*}{4.0} & Euler Solver & 91.62$\pm$0.34 & 93.21$\pm$0.43\\
                             & RK4 Solver &91.00$\pm$1.31  & 97.67$\pm$0.22\\ 
         \midrule
        \multirow{2}{*}{5.0} & Euler Solver & 91.16$\pm$0.67 & 93.11$\pm$2.36\\
                             & RK4 Solver & 90.26$\pm$2.16 & 97.10$\pm$2.02\\ 
      
        \bottomrule
          
    \end{tabular}
    \caption{Node classification accuracy under different integration times of CDE. The ODE solver is used with step size $\tau=1$. We fixed all other hyperparameters in the model except the integration time $T$.  Unlike the traditional GNNs, the neural diffusion models do not have explicit layers but we may consider the steps taken to solve an ODE as the implicit layers. Hence, the number of implicit layers in neural diffusion models is $T/\tau$ for fixed step-size ODE solvers.}
    \label{tab:abatime}
\end{table}

\begin{table*}[!htp]\small
\centering
 \resizebox{0.95\textwidth}{!}{\begin{tabular}{ccccccccc}
\toprule
Method  & Wiki-cooc & Roman-empire & Amazon-ratings & Minesweeper & Workers & Questions  \\
\midrule
GRAND-LAP &  91.58$\pm$0.37  &  69.24$\pm$0.53  &  48.99$\pm$0.35  &  73.25$\pm$0.99  &  75.59$\pm$0.86  & 68.54$\pm$1.07  \\
GRAND-GAT &  92.03$\pm$0.46  &  71.60$\pm$0.58  &  45.05$\pm$0.65  &   76.67$\pm$0.98  &  75.33$\pm$0.84  & 70.67$\pm$1.28  \\
GRAND-TRANS &  91.86$\pm$0.27  &  71.18$\pm$0.56  &  45.20$\pm$0.52  &  75.40$\pm$1.36  &  75.13$\pm$0.65  & 69.14$\pm$0.97  \\
GraphBel &  90.30$\pm$0.50  &  69.47$\pm$0.37  &  43.63$\pm$0.42  &  76.51$\pm$1.03  &  73.02$\pm$0.92  & 70.79$\pm$0.99  \\
\midrule
CDE-GRAND-LAP &  98.00$\pm$0.20  &  90.58$\pm$0.49  &  47.43$\pm$0.53  &  90.06$\pm$0.60  &  80.45$\pm$0.97  & 73.78$\pm$1.46  \\
CDE-GRAND-GAT &  97.99$\pm$0.38  &  91.64$\pm$0.28  &  47.63$\pm$0.43  &  95.50$\pm$5.23  &  80.70$\pm$1.04  & 75.17$\pm$0.99  \\

CDE-GRAND-TRANS &    98.04$\pm$0.35  &  91.55$\pm$0.23  &  46.91$\pm$0.87  &  91.38$\pm$1.92  &  81.58$\pm$0.98  & 72.92$\pm$1.54  \\
CDE-GraphBel &  97.79$\pm$0.40  &  85.39$\pm$0.46  &  45.22$\pm$0.60  &  90.79$\pm$0.48  &  81.30$\pm$0.43  & 72.11$\pm$1.31  \\
\bottomrule
\end{tabular}}
\caption{Node classification accuracy(\%). We evaluate the performance of the vanilla diffusion variants and the inclusion of CDE convection.}
\label{tab:abadiffusion}
\end{table*}

\vspace{0.1cm}

\begin{figure}[ht]
    \centering
    \includegraphics[width=\linewidth]{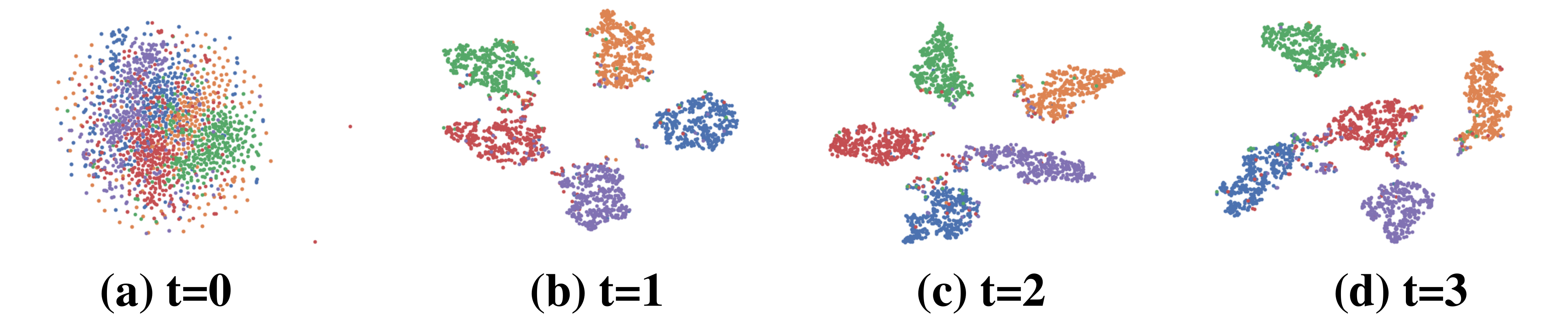}
    \caption{T-sne visualization of the feature representation from CDE-based neural diffusion models at different integration time $t$ on the synthetic dataset described in \cref{subs:Syn} . From left to right: $t=0$, $t=1$, $t=2$ and $t=3$}
    \label{fig:my_label}
\end{figure}

\subsection{Visulization}
The visualization of the node features from our CDE model at different integration time $t$ is shown in \cref{fig:my_label}.
From \cref{fig:my_label}, we observe that the heterophilic graph dataset are classified into different classes by using our CDE models obviously. \\

\subsection{Ablation Studies}
\subsubsection{Integration Time}
We can see from \cref{tab:abatime} that as the integration time of our proposed CDE model increases, the node classification accuracy improves on the Roman-empire and Minesweeper datasets. However, there is a point at which the predicted classification results reach a saturation point. Therefore, it is important to consider the balance between accuracy and computation time when selecting the appropriate integration time for CDE, as a larger integration time results in increased computation time.

\subsubsection{PDE Solvers}
There are various numerical methods available for solving nonlinear diffusion equations. In this work, we primarily utilize explicit solvers provided in \cite{chen2018neural} for our model. These solvers can be divided into single-step and multi-step schemes, where the latter involves the use of multiple function evaluations at different time steps to compute the next iteration. This results in multi-hop neighboring information aggregation in multi-step explicit solvers. In this work, we use a multi-step solver named RK4 and a single-step explicit solver named Euler. As seen in \cref{tab:abatime}, RK4 outperforms Euler, particularly on the Minesweeper dataset. This is likely due to the higher-order neighborhood aggregation used in RK4, which has been shown to be effective for heterophilic graphs \cite{zhuyanhei:designs2020}. However, the use of an RK4 solver typically results in higher computational complexity. In order to balance performance and computational efficiency, we choose to use the Euler solver for the experiments reported in
\cref{tab:noderesults}, even though better performance may be achieved with the RK4 solver.

\begin{table}[!htp]\small
\centering
\begin{tabular}{ccc}
\toprule
Method  & Training Time (ms) & Inference Time (ms)  \\
\midrule

GRAND &  9.55  & 4.66  \\
GraphBel & 12.55  & 6.69   \\

\midrule
CDE-GRAND & 10.43   & 4.72  \\
CDE-GraphBel & 13.91  & 6.81  \\

\bottomrule
\end{tabular}
\caption{Training time and inference time of models on the Cora dataset using Euler solver. Neural PDE models are based on the Euler solver with an integral time of 1.0 and a hidden dim of 64. The training and inference time refers to the average time for 100 rounds.}
\label{tab:runtime}
\end{table}

\subsubsection{Diffusion Functions}\label{sec.variousdiff_f}
In this section, we show the node classification performance under different diffusion functions compared to the vanilla diffusion variants defined in \cite{chamrowgor:grand2021,SonKanWan:C22}. 
The learnable attention matrix  $\mathbf{A}(\bX(t))= \left(a\left(\bx_i(t),  \bx_j(t)\right)\right)$ in \cref{eq.GRAND} can be formulated with three versions as used in \cite{chamrowgor:grand2021}.

\tb{GRAND-TRANS} uses the scaled dot product attention from the Transformer model \cite{vaswani2017attention}:
\begin{align}
    a(\bx_i,\bx_j) = \mathrm{softmax} \left(\frac{(\mathbf{W}_K \bx_{i} )^\top \mathbf{W}_Q \bx_{j}}{d_k}\right),
\end{align}
where $\mathbf{W}_K$ and $\mathbf{W}_Q$ are learned matrices, and $d_k$ is a hyperparameter determining the dimension of $W_k$. 

\tb{GRAND-GAT} uses the  attention in GAT model \cite{velickovic2017gat}:
\begin{align}\label{eq:bahdanau_attention}
a(\bx_i,\bx_j) = \frac{\exp \left(\mathrm{LeakyReLU} \left(\mathbf{a}\T\left[\bW \bx_{i} \| \bW \bx_{j}\right]\right)\right)}{\sum_{k \in \mathcal{N}_{i}} \exp \left(\mathrm{LeakyReLU}\left(\mathbf{a}\T\left[\bW \bx_{i} \| W \bx_{k}\right]\right)\right)}, 
\end{align}

where $\bW$ and $\ba$ are learned and $\|$ is the concatenation operator.

\tb{GRAND-LAP} models the linear diffusion process, where the $\bA(t)$ a constant in the integral, i.e., $\mathbf{A}(\bX(t)) = \mathbf{A}$ for a constant $\mathbf{A}$.

 \tb{GraphBel} is inspired by Beltrami diffusion \cite{sochenTIP1998}. We refer readers to \cite{SonKanWan:C22} for its diffusion formulation.

The results in table \ref{tab:abadiffusion} show that our CDE framework significantly improves the node classification performance of various diffusion-based GNN models on different datasets. 
The improvement is particularly noteworthy on the Roman-empire and Minesweeper datasets, with increases of over or near 15\%.
This highlights the effectiveness of incorporating the convection term into traditional diffusion-based GNNs.

\subsubsection{Time Complexity}

From \cref{tab:runtime}, we observe that introducing CDE for the GNNs does not increase the training and inference time too much.
Specifically, around 10\% is increased for training time, while, over 1\% is increased for inference time. 
As a result, it is acceptable to use CDE for GNNs to improve the performance w.r.t running complexity.

\section{Conclusion}
In order to handle the heterophilic graph efficiently, we introduce the convection-diffusion equation into GNNs to represent the information flow w.r.t nodes. Based on this framework, the homophilic information ``diffusion''  and the heterophilic information ``convection'' are combined for the graph representation. The extensive experiments show that our framework achieves competitive performance on node classification tasks for heterophilic graphs, compared to the state-of-the-art methods. Therefore, the convection-diffusion equation for the GNNs is beneficial to heterophilic graph learning.

\appendix

\section*{Ethical Statement}
As for the limitations, we though have performed our method on several heterophilic graph datasets for the node classification task, it is necessary to further verify the performance on other challenging tasks related to the heterophilic graph representation, such as users' preference recommendation on social networks. Moreover, we have also attempted to investigate the adversarial and defense for our method on heterophilic graph datasets, which is also related to practical applications with robustness requirements. 
As for the ethical statement, our work just designs a heterophilic graph learning approach to handle open source datasets without ethical problems.  

\section*{Acknowledgments}
This research is supported by A*STAR under its RIE2020 Advanced Manufacturing and Engineering (AME) Industry Alignment Fund – Pre Positioning (IAF-PP) (Grant No. A19D6a0053) and the National Research Foundation, Singapore and Infocomm Media Development Authority under its Future Communications Research and Development Programme. The computational work for this article was partially performed on resources of the National Supercomputing Centre, Singapore (https://www.nscc.sg).

\newpage

\bibliographystyle{named}
\bibliography{bib/IEEEabrv,ijcai23}

\begin{thebibliography}{}

\bibitem[\protect\citeauthoryear{Bo \bgroup \em et al.\egroup
  }{2021}]{boxiashi:beyondlow2021}
Deyu Bo, Xiao Wang, Chuan Shi, and Huawei Shen.
\newblock Beyond low-frequency information in graph convolutional networks.
\newblock In {\em Proceedings of the AAAI Conference on Artificial
  Intelligence}, volume~35, pages 3950--3957, 2021.

\bibitem[\protect\citeauthoryear{Bodnar \bgroup \em et al.\egroup
  }{2022}]{crifraben:sheaf2022}
Cristian Bodnar, Francesco~Di Giovanni, Benjamin~Paul Chamberlain, Pietro
  Li{\`o}, and Michael~M. Bronstein.
\newblock Neural sheaf diffusion: A topological perspective on heterophily and
  oversmoothing in {GNN}s.
\newblock In {\em Advances in Neural Information Processing Systems}, 2022.

\bibitem[\protect\citeauthoryear{Chamberlain \bgroup \em et al.\egroup
  }{2021a}]{chamrowgor:grand2021}
Ben Chamberlain, James Rowbottom, Maria~I Gorinova, Michael Bronstein, Stefan
  Webb, and Emanuele Rossi.
\newblock Grand: Graph neural diffusion.
\newblock In {\em International Conference on Machine Learning}, pages
  1407--1418. PMLR, 2021.

\bibitem[\protect\citeauthoryear{Chamberlain \bgroup \em et al.\egroup
  }{2021b}]{charoweyn:blend2021}
Benjamin Chamberlain, James Rowbottom, Davide Eynard, Francesco Di~Giovanni,
  Xiaowen Dong, and Michael Bronstein.
\newblock Beltrami flow and neural diffusion on graphs.
\newblock {\em Advances in Neural Information Processing Systems},
  34:1594--1609, 2021.

\bibitem[\protect\citeauthoryear{Chamberlain \bgroup \em et al.\egroup
  }{2021c}]{chamberlain2021grand}
Benjamin~Paul Chamberlain, James Rowbottom, Maria Goronova, Stefan Webb,
  Emanuele Rossi, and Michael~M Bronstein.
\newblock Grand: Graph neural diffusion.
\newblock In {\em Proc. Int. Conf. Mach. Learn.}, 2021.

\bibitem[\protect\citeauthoryear{Chanpuriya and
  Musco}{2022}]{chansudcam:simplified2022}
Sudhanshu Chanpuriya and Cameron Musco.
\newblock Simplified graph convolution with heterophily.
\newblock {\em arXiv preprint arXiv:2202.04139}, 2022.

\bibitem[\protect\citeauthoryear{Chen \bgroup \em et al.\egroup
  }{2018}]{chen2018neural}
Ricky~TQ Chen, Yulia Rubanova, Jesse Bettencourt, and David Duvenaud.
\newblock Neural ordinary differential equations.
\newblock In {\em Proc. Advances Neural Inf. Process. Syst.}, 2018.

\bibitem[\protect\citeauthoryear{Chien \bgroup \em et al.\egroup
  }{2021}]{chipenli:gprgnn2021}
Eli Chien, Jianhao Peng, Pan Li, and Olgica Milenkovic.
\newblock Adaptive universal generalized pagerank graph neural network.
\newblock In {\em International Conference on Learning Representations}, 2021.

\bibitem[\protect\citeauthoryear{Du \bgroup \em et al.\egroup
  }{2022}]{dushifu:gbkgnn2022}
Lun Du, Xiaozhou Shi, Qiang Fu, Xiaojun Ma, Hengyu Liu, Shi Han, and Dongmei
  Zhang.
\newblock Gbk-gnn: Gated bi-kernel graph neural networks for modeling both
  homophily and heterophily.
\newblock In {\em Proceedings of the ACM Web Conference 2022}, pages
  1550--1558, 2022.

\bibitem[\protect\citeauthoryear{Grigoryan}{2009}]{grigoryan2009heat}
Alexander Grigoryan.
\newblock {\em Heat kernel and analysis on manifolds}.
\newblock American Mathematical Soc., Providence, 2009.

\bibitem[\protect\citeauthoryear{Hamilton \bgroup \em et al.\egroup
  }{2017}]{hamilton2017sage}
William~L. Hamilton, Rex Ying, and Jure Leskovec.
\newblock Inductive representation learning on large graphs.
\newblock In {\em Proc. Advances Neural Inf. Process. Syst.}, 2017.

\bibitem[\protect\citeauthoryear{He \bgroup \em et al.\egroup
  }{2016}]{he2016resnet}
Kaiming He, Xiangyu Zhang, Shaoqing Ren, and Jian Sun.
\newblock Deep residual learning for image recognition.
\newblock In {\em Proceedings of the IEEE conference on computer vision and
  pattern recognition}, pages 770--778, 2016.

\bibitem[\protect\citeauthoryear{Ji \bgroup \em et al.\egroup
  }{2023}]{JiLeeMen:labelgnn}
Feng Ji, See~Hian Lee, Hanyang Meng, Kai Zhao, Jielong Yang, and Wee~Peng Tay.
\newblock Leveraging label non-uniformity for node classification in graph
  neural networks.
\newblock In {\em Proc. International Conference on Machine Learning}, Haiwaii,
  USA, Jul. 2023.

\bibitem[\protect\citeauthoryear{Kang \bgroup \em et al.\egroup
  }{2021}]{KanSonDinTay:stableode}
Qiyu Kang, Yang Song, Qinxu Ding, and Wee~Peng Tay.
\newblock Stable neural {ODE} with {Lyapunov}-stable equilibrium points for
  defending against adversarial attacks.
\newblock In {\em Advances in Neural Information Processing Systems (NeurIPS)},
  virtual, Dec. 2021.

\bibitem[\protect\citeauthoryear{Kang \bgroup \em et al.\egroup
  }{2023}]{KanZhaSon:hamgnn}
Qiyu Kang, Kai Zhao, Yang Song, Sijie Wang, and Wee~Peng Tay.
\newblock Node embedding from neural {Hamiltonian} orbits in graph neural
  networks.
\newblock In {\em Proc. International Conference on Machine Learning}, Haiwaii,
  USA, Jul. 2023.

\bibitem[\protect\citeauthoryear{Kingma and Ba}{2014}]{kingma2014adam}
Diederik~P Kingma and Jimmy Ba.
\newblock Adam: A method for stochastic optimization.
\newblock {\em arXiv preprint arXiv:1412.6980}, 2014.

\bibitem[\protect\citeauthoryear{Kipf and Welling}{2016}]{kipf2016vgae}
Thomas~N. Kipf and Max Welling.
\newblock Variational graph auto-encoders.
\newblock In {\em Proc. Advances Neural Inf. Process. Syst. Workshop}, 2016.

\bibitem[\protect\citeauthoryear{Kipf and Welling}{2017}]{kipfgcn}
Thomas~N Kipf and Max Welling.
\newblock Semi-supervised classification with graph convolutional networks.
\newblock In {\em Proc. Int. Conf. Learn. Representations}, pages 1--14, 2017.

\bibitem[\protect\citeauthoryear{Klicpera \bgroup \em et al.\egroup
  }{2019}]{Klibojste:appnp}
Johannes Klicpera, Aleksandar Bojchevski, and Stephan G{\"{u}}nnemann.
\newblock Predict then propagate: Graph neural networks meet personalized
  pagerank.
\newblock In {\em Proc. Int. Conf. Learning Representations}, 2019.

\bibitem[\protect\citeauthoryear{Lee \bgroup \em et al.\egroup
  }{2022}]{LeeJiTay:simgat}
See~Hian Lee, Feng Ji, and Wee~Peng Tay.
\newblock {SGAT}: Simplicial graph attention network.
\newblock In {\em Proc. International Joint Conference on Artificial
  Intelligence (IJCAI)}, Vienna, Austria, Jul. 2022.

\bibitem[\protect\citeauthoryear{Li \bgroup \em et al.\egroup
  }{2022}]{liyuche:find2022}
Xiang Li, Renyu Zhu, Yao Cheng, Caihua Shan, Siqiang Luo, Dongsheng Li, and
  Weining Qian.
\newblock Finding global homophily in graph neural networks when meeting
  heterophily.
\newblock {\em arXiv preprint arXiv:2205.07308}, 2022.

\bibitem[\protect\citeauthoryear{Luan \bgroup \em et al.\egroup
  }{2022}]{luahualu:revisit2022}
Sitao Luan, Chenqing Hua, Qincheng Lu, Jiaqi Zhu, Mingde Zhao, Shuyuan Zhang,
  Xiao-Wen Chang, and Doina Precup.
\newblock Revisiting heterophily for graph neural networks.
\newblock {\em arXiv preprint arXiv:2210.07606}, 2022.

\bibitem[\protect\citeauthoryear{Ma \bgroup \em et al.\egroup
  }{2021}]{maliusha:homo2021}
Yao Ma, Xiaorui Liu, Neil Shah, and Jiliang Tang.
\newblock Is homophily a necessity for graph neural networks?
\newblock In {\em International Conference on Learning Representations}, 2021.

\bibitem[\protect\citeauthoryear{Pei \bgroup \em et al.\egroup
  }{2019}]{peiweicha:geomgcn2019}
Hongbin Pei, Bingzhe Wei, Kevin Chen-Chuan Chang, Yu~Lei, and Bo~Yang.
\newblock Geom-gcn: Geometric graph convolutional networks.
\newblock In {\em International Conference on Learning Representations}, 2019.

\bibitem[\protect\citeauthoryear{Platonov \bgroup \em et al.\egroup
  }{2022}]{plakuzbab:adjusted2022}
Oleg Platonov, Denis Kuznedelev, Artem Babenko, and Liudmila Prokhorenkova.
\newblock Characterizing graph datasets for node classification: Beyond
  homophily-heterophily dichotomy.
\newblock {\em arXiv preprint arXiv:2209.06177}, 2022.

\bibitem[\protect\citeauthoryear{Platonov \bgroup \em et al.\egroup
  }{2023}]{Data_paper}
Oleg Platonov, Denis Kuznedelev, Michael Diskin, Artem Babenko, and Liudmila
  Prokhorenkova.
\newblock A critical look at evaluation of gnns under heterophily: Are we
  really making progress?
\newblock In {\em The Eleventh International Conference on Learning
  Representations}, 2023.

\bibitem[\protect\citeauthoryear{Rozemberczki \bgroup \em et al.\egroup
  }{2021}]{rozcarrik:multi2021}
Benedek Rozemberczki, Carl Allen, and Rik Sarkar.
\newblock Multi-scale attributed node embedding.
\newblock {\em Journal of Complex Networks}, 9(2):cnab014, 2021.

\bibitem[\protect\citeauthoryear{Rusch \bgroup \em et al.\egroup
  }{2022}]{ruscharow:graphcon2022}
T~Konstantin Rusch, Ben Chamberlain, James Rowbottom, Siddhartha Mishra, and
  Michael Bronstein.
\newblock Graph-coupled oscillator networks.
\newblock In {\em International Conference on Machine Learning}, pages
  18888--18909. PMLR, 2022.

\bibitem[\protect\citeauthoryear{Sochen \bgroup \em et al.\egroup
  }{1998}]{sochenTIP1998}
N.~Sochen, R.~Kimmel, and R.~Malladi.
\newblock A general framework for low level vision.
\newblock {\em {IEEE} Trans. Image Process.}, 7(3):310--318, 1998.

\bibitem[\protect\citeauthoryear{Song \bgroup \em et al.\egroup
  }{2021}]{SonKanTay:errorcorr}
Yang Song, Qiyu Kang, and Wee~Peng Tay.
\newblock Error-correcting output codes with ensemble diversity for robust
  learning in neural networks.
\newblock In {\em Proc. AAAI Conference on Artificial Intelligence}, virtual,
  Feb. 2021.

\bibitem[\protect\citeauthoryear{Song \bgroup \em et al.\egroup
  }{2022}]{SonKanWan:C22}
Yang Song, Qiyu Kang, Sijie Wang, Kai Zhao, and Wee~Peng Tay.
\newblock On the robustness of graph neural diffusion to topology
  perturbations.
\newblock In {\em Advances in Neural Information Processing Systems (NeurIPS)},
  New Orleans, USA, Nov. 2022.

\bibitem[\protect\citeauthoryear{Sun \bgroup \em et al.\egroup
  }{2022}]{sunhapyan:beyond2022}
Yifei Sun, Haoran Deng, Yang Yang, Chunping Wang, Jiarong Xu, Renhong Huang,
  Linfeng Cao, Yang Wang, and Lei Chen.
\newblock Beyond homophily: Structure-aware path aggregation graph neural
  network.
\newblock In {\em Proceedings of the Thirty-First International Joint
  Conference on Artificial Intelligence, {IJCAI-22}}, 2022.

\bibitem[\protect\citeauthoryear{Vaswani \bgroup \em et al.\egroup
  }{2017}]{vaswani2017attention}
Ashish Vaswani, Noam Shazeer, Niki Parmar, Jakob Uszkoreit, Llion Jones,
  Aidan~N Gomez, {\L}ukasz Kaiser, and Illia Polosukhin.
\newblock Attention is all you need.
\newblock {\em Advances in neural information processing systems}, 30, 2017.

\bibitem[\protect\citeauthoryear{Veli{\v{c}}kovi{\'c} \bgroup \em et al.\egroup
  }{2018}]{velickovic2017gat}
Petar Veli{\v{c}}kovi{\'c}, Guillem Cucurull, Arantxa Casanova, Adriana Romero,
  Pietro Lio, and Yoshua Bengio.
\newblock Graph attention networks.
\newblock In {\em Proc. Int. Conf. Learn. Representations}, pages 1--12, 2018.

\bibitem[\protect\citeauthoryear{Wang \bgroup \em et al.\egroup
  }{2022}]{wanjinrui:powerful}
Tao Wang, Di~Jin, Rui Wang, Dongxiao He, and Yuxiao Huang.
\newblock Powerful graph convolutional networks with adaptive propagation
  mechanism for homophily and heterophily.
\newblock In {\em Proceedings of the AAAI Conference on Artificial
  Intelligence}, pages 4210--4218, 2022.

\bibitem[\protect\citeauthoryear{Wang \bgroup \em et al.\egroup
  }{2023a}]{WanKanShe:robustloc}
Sijie Wang, Qiyu Kang, Rui She, Wee~Peng Tay, Andreas Hartmannsgruber, and
  Diego~Navarro Navarro.
\newblock {RobustLoc}: {Robust} camera pose regression in challenging driving
  environments.
\newblock In {\em Proc. AAAI Conference on Artificial Intelligence},
  Washington, DC, Feb. 2023.

\bibitem[\protect\citeauthoryear{Wang \bgroup \em et al.\egroup
  }{2023b}]{wang2023acmp}
Yuelin Wang, Kai Yi, Xinliang Liu, Yu~Guang Wang, and Shi Jin.
\newblock {ACMP}: Allen-cahn message passing with attractive and repulsive
  forces for graph neural networks.
\newblock In {\em The Eleventh International Conference on Learning
  Representations}, 2023.

\bibitem[\protect\citeauthoryear{Wu \bgroup \em et al.\egroup
  }{2019}]{wuaouzha:sgc2019}
Felix Wu, Amauri Souza, Tianyi Zhang, Christopher Fifty, Tao Yu, and Kilian
  Weinberger.
\newblock Simplifying graph convolutional networks.
\newblock In {\em International conference on machine learning}, pages
  6861--6871. PMLR, 2019.

\bibitem[\protect\citeauthoryear{Xu \bgroup \em et al.\egroup
  }{2019}]{xu2018gin}
Keyulu Xu, Weihua Hu, Jure Leskovec, and Stefanie Jegelka.
\newblock How powerful are graph neural networks?
\newblock In {\em Proc. Int. Conf. Learn. Representations}, 2019.

\bibitem[\protect\citeauthoryear{Yang \bgroup \em et al.\egroup
  }{2021}]{tanliliu:diverse2021}
Liang Yang, Mengzhe Li, Liyang Liu, Chuan Wang, Xiaochun Cao, Yuanfang Guo,
  et~al.
\newblock Diverse message passing for attribute with heterophily.
\newblock {\em Advances in Neural Information Processing Systems},
  34:4751--4763, 2021.

\bibitem[\protect\citeauthoryear{Zhu \bgroup \em et al.\egroup
  }{2020}]{zhuyanhei:designs2020}
Jiong Zhu, Yujun Yan, Lingxiao Zhao, Mark Heimann, Leman Akoglu, and Danai
  Koutra.
\newblock Beyond homophily in graph neural networks: Current limitations and
  effective designs.
\newblock {\em Advances in Neural Information Processing Systems},
  33:7793--7804, 2020.

\bibitem[\protect\citeauthoryear{Zhu \bgroup \em et al.\egroup
  }{2021}]{zhurosrao:graphheter}
Jiong Zhu, Ryan~A Rossi, Anup Rao, Tung Mai, Nedim Lipka, Nesreen~K Ahmed, and
  Danai Koutra.
\newblock Graph neural networks with heterophily.
\newblock In {\em Proceedings of the AAAI Conference on Artificial
  Intelligence}, pages 11168--11176, 2021.

\bibitem[\protect\citeauthoryear{Zhu \bgroup \em et al.\egroup
  }{2022}]{zhu2022does}
Jiong Zhu, Junchen Jin, Donald Loveland, Michael~T Schaub, and Danai Koutra.
\newblock How does heterophily impact the robustness of graph neural networks?
  theoretical connections and practical implications.
\newblock In {\em Proceedings of the 28th ACM SIGKDD Conference on Knowledge
  Discovery and Data Mining}, pages 2637--2647, 2022.

\end{thebibliography}


\appendix
\clearpage

\textbf{\LARGE Appendix}
\vspace{0.6cm}

The non-prefixed equation numbers and result references correspond to those in the main paper.  Labels in this supplementary material are prefixed with ``S''. 
 
Summary of the supplementary material:
\begin{enumerate}
    \item In \cref{sec.supp_exi_gra_diff}, we present an overview of existing diffusion-based graph PDE models from literature such as \cite{chamrowgor:grand2021,SonKanWan:C22,crifraben:sheaf2022}.
    \item In \cref{sec.fur}, we offer a detailed elucidation of our model, complementing the information provided in the main paper. 
    \item In \cref{sec.supp_exp_set}, we provide additional information on the experimental setup outlined in \cref{sec.exp} of the main paper, including specific hyperparameter values used for our models on each dataset.
    \item In \cref{sec.supp_more_exp}, we provide more experiments to demonstrate the effectiveness of our method. 
    \item We refer interested readers to our source code for reproducing our results.
\end{enumerate}

\section{Existing Graph Neural Diffusion}\label{sec.supp_exi_gra_diff}
In this section, we present  existing diffusion-based graph PDE models that are not included in the main paper due to space constraints.

\tb{GRAND:} Derived from the heat diffusion equation, GRAND \cite{chamrowgor:grand2021} implements the following dynamical system:
\begin{align}
\frac{{\partial} \mathbf{X}(t)}{{\partial} t} 
=(\mathbf{A}(\mathbf{X}(t))-\mathbf{I}) \mathbf{X}(t) := \overline{\mathbf{A}}(\mathbf{X}(t)) \mathbf{X}(t), \label{eq.GRAND_full}
\end{align} 
with the initial condition $\mathbf{X}(0)$. The matrix $\mathbf{A}(\mathbf{X}(t))= \left(a\left(\mathbf{x}_i(t),  \mathbf{x}_j(t)\right)\right)$ is a learnable attention matrix to describe the structure of the graph, $a(\cdot)$ is a similarity function for vertex pairs, and $\mathbf{I}$ is an identity matrix.

\tb{GraphBel:} Generalizing the Beltrami flow, mean curvature flow and heat flow, a more robust graph neural flow \cite{SonKanWan:C22} is designed as
\begin{align}
\frac{{\partial} \mathbf{X}(t)}{{\partial} t}=(\mathbf{A_S}(\mathbf{X}(t)) \odot \mathbf{B_S}(\mathbf{X}(t))-\Psi(\mathbf{X}(t))) \mathbf{X}(t),  \label{eq.SonKan}
\end{align}
where $\odot$ is the element-wise multiplication. $\mathbf{A_S}(\cdot)$ and $\mathbf{B_S}(\cdot)$ are learnable attention function and normalized vector map, respectively.  
$\mathbf{\Psi}(\mathbf{X}(t))$ is a diagonal matrix in which $\Psi(\mathbf{x}_i, \mathbf{x}_i)=\sum_{\mathbf{x}_j}(\mathbf{A} \odot \mathbf{B})(\mathbf{x}_i, \mathbf{x}_j)$.

\tb{Neural Sheaf Diffusion:}  The paper \cite{crifraben:sheaf2022} models the sheaf diffusion process on graph. The diffusion equation is given by   
\begin{align}\label{eq:sheaf}
    \frac{{\partial} \mathbf{X}(t)}{{\partial} t} = - \sigma \Big(\Delta_{\calF(t)} (\mathbf{I}_n \otimes \mathbf{W}_1) \mathbf{X}(t) \mathbf{W}_2 \Big)
\end{align}
where sheaf Laplacian $\Delta_{\calF(t)}$ is that of a sheaf $(G, \calF(t))$ that can be described by a learnable function of the data $(G, \calF(t)) = g(G, \mathbf{X}(t) ; \theta)$.  $\mathbf{W}_1$ and $\mathbf{W}_2$ are the weight matrice, $\mathbf{I}_n$ is the  identity matrix, , and $\otimes$ denotes the Kronecker product. \\
The model is designed for processing the heterophilic graphs and mitigating the oversmoothing behavior in GNNs. The diffusion \cref{eq:sheaf} can learn the right geometry for node classification tasks by learning the underlying sheaf from the heterophilic or homophilic graph. But they apply the following time-discretized version of the neural sheaf diffusion model when conducting the experiments:
\begin{equation}\label{eq:disc_sheaf}
    \mathbf{X}_{t+1} = (1 + \varepsilon)\mathbf{X}_{t} - \sigma \Big(\Delta_{\calF(t)} (\mathbf{I} \otimes \mathbf{W}_1^t) \mathbf{X}_t \mathbf{W}_2^t \Big)
\end{equation}
where  $\varepsilon  \in [-1, 1]^d$ is an additional learned parameter (i.e. a vector of size $d$).

\begin{table*}[!ht]\small
\centering
 \resizebox{0.95\textwidth}{!}{\begin{tabular}{ccccccccc}
\toprule
Method  & Wiki-cooc & Roman-empire & Amazon-ratings & Minesweeper & Workers & Questions  \\
\midrule
Diag-NSD &  92.06$\pm$0.40  &  77.50$\pm$0.67  &  37.96$\pm$0.20  &  89.59$\pm$0.61  &  79.81$\pm$0.99  & 69.25$\pm$1.15  \\

\midrule

CDE-Diag-NSD &  92.45$\pm$0.67  &  78.99$\pm$0.52  &  40.92$\pm$1.96  &  91.13$\pm$0.80  &  82.81$\pm$0.51  & 73.65$\pm$1.55  \\

\bottomrule
\end{tabular}}
\caption{Node classification accuracy(\%). We evaluate the performance of the vanilla diffusion variants and the inclusion of CDE convection.}
\label{tab:abadiffusion_NSD}
\end{table*}

\section{Further Explanation}\label{sec.fur}
We delve further into the intricacies of our CDE model in the following discussion, where we also compare ACMP with our CDE model.

The ACMP paper \cite{wang2023acmp} and ours have significant differences. The motivation is significantly different. ACMP is inspired by the particle reaction-diffusion process where repulsive and attractive force interactions between particles are considered. On the other hand, our model is inspired by the {convection-diffusion process where the convection process is not from repulsive or attractive forces}. The proposed GNN models consequently generalize GRAND \cite{chamberlain2021grand} from two perspectives. This can be seen from the following GNN formulations: \\
$\bullet$ $\frac{\partial}{\partial t} \mathbf{x}_i(t)
= \sum_{j \in \mathcal{N}_i}a\left(\mathbf{x}_i(t), \mathbf{x}_j(t)\right)\left(\mathbf{x}_j(t)-\mathbf{x}_i(t)\right)+{\sigma(W(\mathbf{x}_j(t) - \mathbf{x}_i(t) ))  \odot \mathbf{x}_i(t)}$  \tb{(Our model - CDE)}\\
$\bullet$ 
$\frac{\partial}{\partial t} \mathbf{x}_i(t)
=\boldsymbol{\alpha} \odot \sum_{j \in \mathcal{N}_i} \left(a\left(\mathbf{x}_i(t), \mathbf{x}_j(t)\right)-\beta\right)\left(\mathbf{x}_j(t)-\mathbf{x}_i(t)\right)+\boldsymbol{\delta} \odot \mathbf{x}_i(t) \odot\left(1-\mathbf{x}_i(t) \odot \mathbf{x}_i(t)\right)$ \tb{(ACMP)}\\
Our model allows a more flexible and adaptive information flow direction using $\sigma(\cdot)$ and $W$.  Furthermore, we dot product this flow with $\bx_i$. This convection term is not the same as the bias in ACMP, which is a weighted sum of $\left(\mathbf{x}_j(t)-\mathbf{x}_i(t)\right)$ and $\mathbf{x}_i(t) \odot\left(1-\mathbf{x}_i(t) \odot \mathbf{x}_i(t)\right)$). (See below for more comparisons.) Our model surpasses ACMP on all the heterophilic datasets, which shows the advantage of our design. 

The velocity term $\sum_{j \in \mathcal{N}_i} \sigma(W(\mathbf{x}_j(t) - \mathbf{x}_i(t))$ in the above blue convection term in CDE learns from the differences between neighbors in heterophily graph datasets. Compared to the repulsive and attractive forces used in ACMP, which have only two directions, $\mathbf{x}_j(t) - \mathbf{x}_i(t)$ and $-(\mathbf{x}_j(t) - \mathbf{x}_i(t))$, our approach is more flexible due to the presence of $W$ and $\sigma(\cdot)$. ACMP only performs a weighted sum over the neighbors' differences using vector-wise attention, while the convection term in CDE further uses the velocity to guide the evolution of $\mathbf{x}_i(t)$ by the dot product in a channel-wise fashion. CDE's superior performance compared to ACMP suggests that updating different feature dimensions with different proportions according to the differences in each dimension is better for heterophily graph datasets. This may be because even if neighboring nodes generally have different features in heterophily datasets, their features in some dimensions may still be close to each other.

\begin{table*}[!htp]\small
\centering
\resizebox{1\textwidth}{!}{
\begin{tabular}{cccccccccccc}
\toprule
Method  & Cornell & Wisconsin & Texas & Film & Chameleon  & Squirrel& Cora & Citeseer & Pubmed  \\
\midrule
MLP &  91.30$\pm$0.70  &  93.87$\pm$3.33  &  92.26$\pm$0.71  &  38.58$\pm$0.25  &  46.72$\pm$0.46  & 31.28$\pm$0.27 & 76.44$\pm$0.30 & 76.25$\pm$0.28 &  86.43$\pm$0.13   \\

\midrule
GCN &  82.46$\pm$3.11  &  75.50$\pm$2.92  &  83.11$\pm$3.20  &  35.51$\pm$0.99  &  64.18$\pm$2.62  & 44.76$\pm$1.39 &  87.78$\pm$0.96 & 81.39$\pm$1.23 & 88.90$\pm$0.32\\
GAT &  76.00$\pm$1.01 & 71.01$\pm$4.66 & 78.87$\pm$0.86 & 35.98$\pm$0.23 & 63.90$\pm$0.46 & 42.72$\pm$0.33 & 76.70$\pm$0.42 & 67.20$\pm$0.46 & 83.28$\pm$0.12  \\
SAGE &   71.41$\pm$1.24 & 64.85$\pm$5.14 & 79.03$\pm$1.20 & 36.37$\pm$0.21 & 62.15$\pm$0.42 & 41.26$\pm$0.26  & 86.58$\pm$0.26 & 78.24$\pm$0.30 & 86.85$\pm$0.11 \\

\midrule
H2GCN & 86.23$\pm$4.71 & 87.50$\pm$1.77 & 85.90$\pm$3.53 & 38.85$\pm$1.17 & 52.30$\pm$0.48 & 30.39$\pm$1.22  & 87.52$\pm$0.61 & 79.97$\pm$0.69 & 87.78$\pm$0.28 \\
GPR-GNN &   91.36$\pm$0.70 & 93.75$\pm$2.37 & 92.92$\pm$0.61 & 39.30$\pm$0.27 & 67.48$\pm$0.40 & 49.93$\pm$0.53 &  79.51$\pm$0.36 & 67.63$\pm$0.38 & 85.07$\pm$0.09   \\
FAGCN &  88.03$\pm$5.60 & 89.75$\pm$6.37 & 88.85$\pm$4.39 & 31.59$\pm$1.37 & 49.47$\pm$2.84 & 42.24$\pm$1.20 & \tb{88.85$\pm$1.36} & \tb{82.37$\pm$1.46} & 89.98$\pm$0.54  \\
ACM-GCN & \tb{94.75$\pm$3.80} & \tb{95.75$\pm$2.03} & \sd{94.92$\pm$2.88} & \tb{41.62$\pm$1.15 } & \tb{69.04$\pm$1.74} & \tb{58.02$\pm$1.86}  & \sd{88.62$\pm$1.22} & \sd{81.68$\pm$0.97} & \tb{90.66$\pm$0.47}  \\
\midrule
GRAND &  89.34$\pm$3.30  &  86.25$\pm$4.58  &  84.26$\pm$3.45  &  38.97$\pm$1.72  &  44.20$\pm$3.27  & 34.06$\pm$1.22 &  86.21$\pm$1.56  &  79.51$\pm$1.55  & 87.22$\pm$1.22 \\
GraphBel &  90.49$\pm$3.09  &  91.13$\pm$4.31  &  88.52$\pm$4.58  &  \sd{40.25$\pm$1.66}  &  51.33$\pm$1.50  & 35.44$\pm$0.96  &  80.02$\pm$1.79  &  79.89$\pm$2.01  & 88.50$\pm$0.53  \\

\midrule
CDE-GRAND &  \sd{92.13$\pm$3.00}  &   \sd{95.13$\pm$3.03}  &  93.61$\pm$4.36  &  39.82$\pm$1.06  &   \sd{68.45$\pm$2.47}  &   \sd{55.04$\pm$1.73}  & 87.19$\pm$1.44 & 80.04$\pm$1.75 &  \sd{90.05$\pm$0.64}  \\ 
CDE-GraphBel &  91.97$\pm$2.79  &  94.50$\pm$2.81  &  \tb{95.57$\pm$4.15}  &  40.08$\pm$1.49  &  64.88$\pm$2.18  &  45.62$\pm$2.12 & 85.55$\pm$1.52  &  78.14$\pm$2.50 & 89.61$\pm$0.88 \\   
\bottomrule
\end{tabular}}
\caption{Node classification accuracy(\%) under 60\%,20\%,20\% random split for training, validation and test set.  The best and the second-best result for each criterion are highlighted in \tb{bold} and \sd{underlined} respectively.  }
\label{tab:randomresults}
\end{table*}

\section{Experiments Setting}\label{sec.supp_exp_set}


\subsection{Implementation Details}

Our CDE model, which incorporates a convection term in the graph diffusion process, is built upon the open-source repository of the GRAND model \cite{chamrowgor:grand2021} available at \url{https://github.com/twitter-research/graph-neural-pde}. The baseline results presented in \cref{tab:noderesults} were taken from the study by Lu et al. \cite{luahualu:revisit2022}.

\subsection{Hyperparameters}
In this section, we provide the hyperparameters in \cref{tab:hyperpara,tab:s2hyperpara} that have been used in our experiments for the model and the training/testing process. 

\begin{table*}[!htp]\small
\centering
\begin{tabular}{c|ccccccc}
    \toprule
    Dataset & Model & lr & weight decay & dropout & hidden dim & time & step size  \\
    \midrule
    \multirow{2}{*}{Texas} & CDE-GRAND & 0.005 & 0.001 & 0.8 & 256 & 0.5 & 0.2\\
    & CDE-GraphBel & 0.005 & 0.0001 & 0.4  & 64 & 0.5 & 0.2\\

    \multirow{2}{*}{Cornell} & CDE-GRAND & 0.005 & 0.01 & 0.6 & 128 & 1.0 &0.2 \\
    & CDE-GraphBel & 0.005 & 0.01 & 0.6 & 128 & 1.5 & 0.2\\

    \multirow{2}{*}{Wisconsin} & CDE-GRAND & 0.005 & 0.01 & 0.6 & 128  & 0.5 &0.2 \\
    & CDE-GraphBel & 0.005 & 0.0001 & 0.4 & 64 & 1.0 & 0.2 \\

     \multirow{2}{*}{Amazon-ratings} & CDE-GRAND & 0.01 & 0.001 & 0.2 & 64 & 1.0 & 0.5 \\
    & CDE-GraphBel & 0.01 & 0.001 & 0.2 & 64 & 1.0 & 1.0 \\

    \multirow{2}{*}{Minesweeper} & CDE-GRAND & 0.01 & 0.001 & 0.2 & 64 & 4.0 & 1.0 \\
    & CDE-GraphBel & 0.01 & 0.001 & 0.2 & 64 & 3.0 & 1.0 \\
     \multirow{2}{*}{Questions} & CDE-GRAND &  0.01 & 0.001 & 0.2 & 64 & 3.0 & 1.0\\
    & CDE-GraphBel & 0.01 & 0.001 & 0.2 & 64 & 1.0 & 1.0 \\

    \multirow{2}{*}{Roman-empire} & CDE-GRAND &  0.01 & 0.001 & 0.2 & 64 & 3.0 & 1.0\\
    & CDE-GraphBel & 0.01 & 0.001 & 0.2 & 64 & 1.0 & 1.0 \\

     \multirow{2}{*}{Wiki-cooc} & CDE-GRAND &  0.01 & 0.001 & 0.2 & 64 & 1.0 & 1.0\\
    & CDE-GraphBel & 0.01 & 0.001 & 0.2 & 64 & 1.0 & 1.0 \\

      \multirow{2}{*}{Workers} & CDE-GRAND &  0.01 & 0.001 & 0.2 & 64 & 1.0 & 1.0\\
    & CDE-GraphBel & 0.01 & 0.001 & 0.2 & 64 & 3.0 & 1.0 \\
    \bottomrule
\end{tabular}

\caption{Hyper-parameters used in \cref{tab:noderesults} }
\label{tab:hyperpara}
\end{table*}

\begin{table*}[!htp]\small
\centering
\begin{tabular}{c|ccccccc}
    \toprule
    Dataset & Model & lr & weight decay & dropout & hidden dim & time & step size  \\
    \midrule
    \multirow{2}{*}{Texas} & CDE-GRAND & 0.005 & 0.01 & 0.0 & 64 & 0.5 & 0.5\\
    & CDE-GraphBel & 0.005 & 0.01 & 0.4  & 64 & 0.5 & 0.5\\

    \multirow{2}{*}{Cornell} & CDE-GRAND & 0.005 & 0.01 & 0.6 & 256 & 1.5 &0.2 \\
    & CDE-GraphBel & 0.005 & 0.01 & 0.4 & 256 & 1.0 & 0.2\\

    \multirow{2}{*}{Wisconsin} & CDE-GRAND & 0.005 & 0.01 & 0.8 & 128  & 1.0 &0.2 \\
    & CDE-GraphBel & 0.005 & 0.01 & 0.8 & 128 & 0.5 & 0.2 \\

     \multirow{2}{*}{Film} & CDE-GRAND & 0.005 & 0.001 & 0.0 & 256 & 1.0 & 1.0 \\
    & CDE-GraphBel & 0.005 & 0.001 & 0.2 & 64 & 1.0 & 1.0 \\

    \multirow{2}{*}{Chameleon} & CDE-GRAND & 0.005 & 0.0001 & 0.2 & 128 & 2.0 & 1.0 \\
    & CDE-GraphBel & 0.005 & 0.0001 & 0.2 & 256 & 1.0 & 1.0 \\
     \multirow{2}{*}{Squirrel} & CDE-GRAND &  0.01 & 0.001 & 0.2 & 64 & 3.0 & 1.0\\
    & CDE-GraphBel & 0.01 & 0.001 & 0.2 & 64 & 1.0 & 1.0 \\

    \multirow{2}{*}{Cora} & CDE-GRAND &  0.01 & 0.001 & 0.6 & 64 & 8.0 & 1.0\\
    & CDE-GraphBel & 0.01 & 0.001 & 0.2 & 64 & 1.0 & 0.5 \\

     \multirow{2}{*}{Citeseer} & CDE-GRAND &  0.01 & 0.01 & 0.6 & 64 & 1.0 & 0.5\\
    & CDE-GraphBel & 0.01 & 0.001 & 0.6 & 64 & 3.0 & 1.0 \\

      \multirow{2}{*}{Pubmed} & CDE-GRAND &  0.01 & 0.0001 & 0.6 & 64 & 1.0 & 0.5\\
    & CDE-GraphBel & 0.01 & 0.0001 & 0.6 & 64 & 1.0 & 0.5\\
    \bottomrule
\end{tabular}

\caption{Hyper-parameters used in \cref{tab:randomresults} }
\label{tab:s2hyperpara}
    
\end{table*}

\begin{table*}[htp]\small
    \centering
     \resizebox{0.95\textwidth}{!}{\begin{tabular}{c|c|c|c|c|c|c|c|c|c|c}
    \toprule
         & CDE-GRAND & CDE-GraphBel & GRAND & GraphBel &Diag-NSD & ACM-GCN & GCN & FAGCN & GBK-GNN &GPRGNN\\
    \midrule
        Num of parameters &  36042 & 36041 & 15245 & 15241 &  9090& 711146   &  6789  & 7047  &  13509 & 6800 \\
        Model Size (MB) & 0.138  & 0.138 & 0.058 & 0.058 & 0.035 & 2.713  & 0.026 & 0.027  & 0.052 & 0.026\\
        Inference Time (ms) & 28.430  & 46.799 & 20.194 & 29.488 & 103.537 & 5.861 & 18.839 & 40.878  & 29.123 & 20.690 \\
        Training Time (ms) & 89.619  & 134.544 & 58.494 & 83.037 & 272.422 & 28.084 & 31.353 & 81.242  & 39.390 & 35.443 \\
    \bottomrule
    \end{tabular}}
    \caption{Model size and computation time on Wiki-cooc dataset}
\label{tab:time_wiki}
\end{table*}

\begin{table*}[htp]\small
    \centering
     \resizebox{0.95\textwidth}{!}{\begin{tabular}{c|c|c|c|c|c|c|c|c|c|c}
    \toprule
         & CDE-GRAND & CDE-GraphBel & GRAND & GraphBel   &Diag-NSD & GCN & FAGCN & GBK-GNN & GPRGNN\\
    \midrule
        Num of parameters &  6807 & 6806 & 5446 & 5446  & 21499  & 19458 & 19716 & 38850 & 19264\\
        Model Size (MB) & 0.026  & 0.026 & 0.021 & 0.021 & 0.082  & 0.074 & 0.075 & 0.148 & 0.074 \\
        Inference Time (ms) & 2.471  & 4.379 & 1.915 & 3.504 & 12.729  & 1.953 & 3.680 & 4.335 & 1.654\\
        Training Time (ms) & 9.259  & 27.213 & 6.474 & 23.376 & 39.960 & 4.097 & 8.258 & 10.454 & 4.035\\
    \bottomrule
    \end{tabular}}
    \caption{Model size and computation time on Questions dataset}
\label{tab:time_question}
\end{table*}

\section{More Experiments}\label{sec.supp_more_exp}


\subsection{Experiments under Random Splits}

In this section, we conduct experiments on popular heterophilic and homophilic graph benchmarks as shown in \cref{sec:datasets}. We follow the same datasets split setting in \url{https://github.com/SitaoLuan/ACM-GNN/blob/89586d13cc72d035fe0c60b6115d4e6cb8c8f5fc/ACM-Pytorch/utils.py#L462}, which is 60\% for the training set, 20\% for validation set and 20\% for the test set.
We conduct experiments to see whether our framework can achieve competitive performance on node classification tasks for both heterophilic and homophilic graphs, compared to the state-of-the-art methods.

On the homophilic graph datasets, Cora, Citeseer, and Pubmed, \tb{our CDE-GRAND and CDE-GraphBel models perform similarly to or slightly better than the vanilla versions of GRAND and GraphBel.} This supports our hypothesis that the inclusion of the convection term does not negatively impact performance on homophilic graph datasets and that the learnable parameter $W$ in \cref{eq.rgdadfa} is able to adapt well to these types of datasets. When compared with other state-of-the-art methods on these homophilic datasets, our CDE model remains competitive, with only 0.61\% worse than the ACM-GCN model on the Pubmed dataset. 

In the main paper, we do not report the model performance for some heterophily datasets like Squirrel and Chameleon that have been widely used in the literature since that a recent paper \cite{Data_paper} has pointed out a train-test data leakage problem in the Squirrel and Chameleon datasets collected by \cite{rozcarrik:multi2021}, leading to questions on model performance on these two datasets. In this supplementary material, we however still include those datasets and report the experiment results in \cref{tab:randomresults} in case some readers are interested. We observe that on the six traditional heterophilic graph datasets, our model remains competitive and achieves the best on the Texas dataset.

We thus conclude that our CDE framework is suitable for both homophilic and heterophilic graph datasets.

\subsection{Experiments of CDE-Diag-NSD}
In \cref{tab:noderesults} in the main paper, we miss the CDE-Diag-NSD model which adds our convection term to the Diag-NSD diffusion model. In this supplementary material, we report its performance in \cref{tab:abadiffusion_NSD}. We observe that our CDE framework still improves the node classification performance of the Diag-NSD diffusion model consistently, even though Diag-NSD is a specialized diffusion model designed to deal with heterophilic datasets. This observation further validates the effectiveness of our CDE framework.

\subsection{Model size and Computation time}

The computational time and model size of our CDE model, as well as other baseline models, are presented in \cref{tab:time_wiki} and \cref{tab:time_question}. As observed, for these two large datasets, our CDE-GRAND and CDE-GraphBel models exhibit a slightly higher inference time compared with their base models, GRAND and GraphBel, respectively. However, this increase in computational time is marginal, demonstrating the effectiveness of our model. Despite having a similar computational cost as the basic neural diffusion model, our approach achieves significantly improved performance on heterophilic datasets.



\end{document}